\documentclass[11pt]{airesearch}
\usepackage[T1]{fontenc}
\usepackage{lmodern}

\usepackage{fullpage}
\usepackage{wrapfig}
\usepackage{graphicx}
\usepackage{enumitem}
\usepackage{tikz}
\usetikzlibrary{positioning,matrix}
\usepackage{amsmath,amssymb,amsthm}
\usepackage{appendix}
\usepackage{chngcntr}
\usepackage{titletoc}
\usepackage{hyperref}
\usepackage{cleveref}
\setlist[itemize]{leftmargin=*,topsep=1pt,itemsep=1pt}

\DeclareMathOperator{\colim}{colim} 
\newcommand{\E}{\mathcal{E}} 
\newcommand{\J}{\mathcal{J}} 
\newcommand{\C}{\mathcal{C}} 
\newcommand{\Set}{\mathrm{Set}} 
\newcommand{\Sub}{\mathrm{Sub}}
\newcommand{\op}{\mathrm{op}}
\newcommand{\Pred}{\mathrm{Pred}}
\newcommand{\Lang}{\mathrm{LM}} 
\newcommand{\Vocab}{\mathcal{V}} 

\usepackage{algorithm}
\usepackage{algpseudocode}

\usepackage{fancyhdr}
\newcommand{\projectpage}[1]{%
\begin{center}
\small
\vspace{-1.25ex}
\texttt{Project Page}: \url{#1}
\end{center}
}

\ifdefined\final
\usepackage[disable]{todonotes}
\else
\usepackage[textsize=tiny]{todonotes}
\fi

\title{\huge \bfseries \sffamily On the Diagram of Thought}
\author{
\bf \sffamily Yifan Zhang$^{1}$~~~~Yang Yuan$^{1,2}$~~~~Andrew Chi-Chih Yao$^{1,2}$$^{\dagger}$ \\[1.5mm]
$^1$IIIS, Tsinghua University~~~~$^2$Shanghai Qi Zhi Institute \\[0.5mm]
}
\date{}

\begin{document}
\maketitle

\begin{abstract}
Large Language Models (LLMs) excel at many tasks but often falter on complex problems that require structured, multi-step reasoning. We introduce the Diagram of Thought (DoT), a framework that enables a single LLM to build and navigate a mental map of its reasoning. Instead of thinking in a straight line, the model constructs a dynamic diagram of ideas, where it can propose different lines of thought, critique its own steps, and synthesize validated insights into a final conclusion. This process is controller-light: it does not require an external search algorithm or planner, but it does use a deterministic online validator for grammar-constrained typed traces, register constraints, and optional solver checks.
To clarify the reliability target of this process, we ground DoT in a mathematical framework from category theory. We interpret accepted typed reasoning records as diagrams in a slice topos and model synthesis of the selected proposer subdiagram as a finite limit. In the predicate fragment, this same object is equivalently a variance-reversed colimit in the opposite information order. The resulting formalism gives an auditable, step-by-step trace of the LLM's typed reasoning and separates semantic guarantees for the typed subtrace from unconstrained natural-language text and uncertified operational edges.
\end{abstract}

\projectpage{https://github.com/diagram-of-thought/diagram-of-thought}

\section{Introduction}
\label{sec:introduction}

Large Language Models (LLMs) \citep{brown2020language, touvron2023llama} have exhibited remarkable proficiency across a spectrum of natural language tasks. However, achieving robust performance on complex reasoning problems that necessitate structured exploration, iterative refinement, backtracking, and self-correction remains a formidable challenge \citep{huang2022towards}. Initial prompting strategies, such as Chain-of-Thought (CoT) \citep{wei2022chain}, encourage step-by-step reasoning by eliciting intermediate steps. While beneficial, the inherent linearity of CoT struggles to capture the dynamic, non-sequential nature of sophisticated problem-solving, which often involves generating parallel hypotheses, critical evaluation, and synthesis, processes ill-suited to a strictly linear progression.

Recognizing these limitations, subsequent research has explored more complex reasoning structures. Frameworks like Tree-of-Thought (ToT) \citep{yao2023tree} utilize tree structures to manage multiple reasoning paths, while Graph-of-Thought (GoT) \citep{besta2024graph} generalizes this to arbitrary graphs. Other approaches, such as Cumulative Reasoning (CR) \citep{zhang2023cumulative}, leverage multi-agent paradigms with specialized roles. Despite their advancements, these methods frequently require external controllers or multi-component pipelines. In contrast, DoT retains a single-model setting while making the operational constraints explicit and checkable via grammar-constrained masking with register constraints and typed validation.

In this paper, we introduce the Diagram of Thought (DoT) framework, which internalizes complex, iterative reasoning within a single auto-regressive language model, guided by learned role tokens and a typed serialization enforced by an online validator with finite control over record kinds and register/solver checks. DoT conceptualizes the reasoning process as the construction of a Directed Acyclic Graph (DAG), as in Figure \ref{fig:diagram_of_thought}. Nodes represent propositions, critiques, refinements, or verified statements, while edges capture logical or procedural dependencies. The model explores alternatives, responds to critiques, and consolidates validated content toward a conclusion.

A cornerstone of the DoT framework is its operationalization through role-specific tokens (e.g., \texttt{<proposer>}, \texttt{<critic>}, \texttt{<summarizer>}). By learning to predict and condition on these tokens, the model transitions between cognitive roles, generating hypotheses, evaluating them, refining based on feedback, and synthesizing results, within standard auto-regressive decoding. Tokens are training/decoding aids; the recorded \texttt{@node} role is the validator's source of truth. This unifies the entire reasoning process inside one model while keeping the interface auditable. All formal guarantees below are scoped to the typed subtrace accepted by $V$; untyped/free text is semantically inert for $\Phi$ and may diverge in natural language from the typed content.

Crucially, to ensure logical consistency and provide a principled aggregation mechanism, we establish a mathematical foundation for DoT using Topos Theory \citep{maclane2012sheaves, johnstone2002sketches, lambek1988introduction}. This framework allows us to model reasoning steps as formal objects and their synthesis as a universal construction. Specifically, we fix a presheaf topos $\E=\Set^{\C^{\mathrm{op}}}$ and a semantic object $S\in \E$. A typed proposer node is interpreted as an object of the slice $(\E/S)$ (a map $X\to S$); in the predicate instantiation used for most intuitions, this map is a monomorphism $P\hookrightarrow S$ and thus corresponds to a subobject. Synthesis by the \texttt{<summarizer>} role is modeled as a finite limit of the selected proposer diagram in the slice. In the predicate case this is a meet in $\Sub(S)$, equivalently the corresponding colimit after reversing variance into the opposite information order $\Pred(S)=\Sub(S)^{\op}$. Operational dependency edges contribute to provenance; they become semantic arrows only when accompanied by accepted typed certificates. Reflection along a Lawvere--Tierney topology captures validation; for a finite family of already validated ($c$-closed) predicates, the outer closure is redundant because the associated closure is a nucleus. For general slice objects that are not monomorphisms, validation is expressed by applying the induced slice sheafification functor; the phrase ``$c$-closed'' is reserved for subobjects.

Our main contributions are therefore:

\begin{enumerate}[itemsep=1pt, topsep=1pt, partopsep=1pt, parsep=1pt, leftmargin=*]
\item We propose the Diagram of Thought (DoT), a single-model framework for DAG-structured proposals, critiques, and summaries, with a deterministic record serialization whose regular core is supplemented by register and solver checks, together with a finite-control online validator that enables auditable extraction.
\item We develop a categorical semantics for DoT in the slice topos $(\E/S)$ and show that synthesis of a selected finite validated proposer diagram is a finite limit in the slice. In the predicate/mono fragment this is the meet of the selected validated subobjects, equivalently a variance-reversed information-order colimit; in the general-arrow fragment, left-exact slice sheafification reflects this finite limit into the validated/sheaf slice.
\item We define a total and deterministic extraction map from LLM-generated traces to formal diagrams, enabled by a well-formedness discipline (BNF grammar, operational rules) and constrained decoding, ensuring that typed reasoning traces are auditable and structurally sound.
\end{enumerate}

\begin{figure}[ht!]
\centering
\resizebox{0.6\textwidth}{!}{
\begin{tikzpicture}[
node distance=1.5cm and 2.0cm,
proposer_valid/.style={circle, draw=blue!60, fill=blue!20, thick, minimum size=12mm, align=center},
proposer_invalid/.style={circle, draw=blue!60, dashed, thick, minimum size=12mm, align=center},
critic/.style={rectangle, draw=red!60, fill=red!20, thick, minimum size=12mm, align=center},
summarizer/.style={ellipse, draw=green!60, fill=green!20, thick, minimum size=12mm, align=center},
edge_style/.style={draw=black, thick, ->, >=latex},
edge_style2/.style={draw=blue!40, thick, ->, >=latex},
label_style/.style={sloped, anchor=south, text=black, font=\small}
]

\node (start) {Problem Statement};

\node[proposer_invalid, below=of start] (p1) {Proposition\\P1};

\node[critic, right=of p1] (c1) {Critique\\C1};

\draw[edge_style] (start) -- (p1);
\draw[edge_style] (p1) -- (c1);

\node[proposer_valid, below=of p1] (p1_refined) {Proposition\\P1'};

\draw[edge_style] (c1) -- node[label_style]{Refine} (p1_refined);

\node[critic, right=of p1_refined] (c2) {Critique\\C2};
\draw[edge_style] (p1_refined) -- (c2);

\node[proposer_valid, below=of p1_refined] (p1_refined_verified) {Proposition\\P1' {\small (Verified)}};

\draw[edge_style, bend right] (c2) to node[label_style]{Verified} (p1_refined_verified);

\node[proposer_valid, right=of p1_refined_verified] (p3) {Proposition\\P3};
\draw[edge_style] (p1_refined_verified) -- (p3);
\draw[edge_style2] (start) -- (p3);

\node[critic, right=of p3] (c3) {Critique\\C3};
\draw[edge_style] (p3) -- (c3);

\node[proposer_valid, below =of p3] (p3v) {Proposition\\P3 {\small(Verified)}};

\draw[edge_style, bend right] (c3) to node[label_style]{Verified} (p3v);

\node[summarizer, below=of p3v] (s1) {Summarization};
\draw[edge_style] (p1_refined_verified) -- (s1);

\draw[edge_style] (p3v) -- (s1);
\draw[edge_style2] (start) -- (s1);

\matrix [draw, fill=white, below, node distance=1cm] at (current bounding box.north east) {
\node [proposer_valid, label=right:Valid Proposition] {}; \\
\node [proposer_invalid, label=right:Invalidated Proposition] {}; \\
\node [critic, label=right:Critique] {}; \\
\node [summarizer, label=right:Summarization] {}; \\
};

\end{tikzpicture}
}
\caption{High-level illustration of the Diagram of Thought (DoT) process. Edges encode dependencies from earlier to later nodes: a critic depends on the proposition it evaluates (proposer $\to$ critic), and the summarizer depends on validated propositions. A single LLM generates the DAG representing proposing (circles), critiquing (rectangles), repairing/verification, and synthesis (ellipse).}
\label{fig:diagram_of_thought}
\end{figure}

\section{Related Work}
\label{sec:related_work}

The pursuit of robust reasoning within Large Language Models (LLMs) has driven considerable research beyond basic input-output functionality. Initial breakthroughs like Chain-of-Thought (CoT) prompting \citep{wei2022chain, kojima2022large} demonstrated that eliciting intermediate reasoning steps significantly improves performance on complex tasks. CoT effectively linearizes reasoning, enhancing transparency but suffering from rigidity; its sequential nature hinders exploration of alternatives or recovery from early errors without restarting. Methods like Self-consistency \citep{wang2022self} mitigate this by sampling multiple reasoning paths and selecting the majority answer, implicitly acknowledging path diversity but lacking explicit refinement mechanisms.

Recognizing the constraints of linearity, subsequent work explored more complex structures. Tree-of-Thought (ToT) \citep{yao2023tree} introduced tree structures where nodes represent partial solutions and edges denote reasoning operators. ToT utilizes search algorithms (e.g., BFS, DFS) guided by heuristic evaluations (often LLM-based) to explore possibilities, enabling systematic search and backtracking. However, ToT generally necessitates an external controller for search management and pruning. Graph-of-Thought (GoT) \citep{besta2024graph} extends this to arbitrary graphs, allowing for more intricate dependency modeling, such as merging reasoning paths, but often requiring more sophisticated external graph management systems.

Collaborative and iterative refinement approaches offer another perspective. Cumulative Reasoning (CR) \citep{zhang2023cumulative} employs multiple LLM instances (or prompts) assigned specific roles (e.g., proposer, verifier), interacting iteratively. While modular, this introduces coordination overhead. Self-Refine \citep{madaan2023self} focuses on iterative improvement where an LLM critiques and refines its own output, though typically applied to the entire output rather than intermediate reasoning steps within a structured process.

From a foundational perspective, \citet{yuan2023power} uses category theory to analyze the inherent capabilities and limitations of LLMs. This work proves that prompt-based tuning is restricted to ``representable'' tasks within the pretext task category, potentially explaining the limitations of simpler methods like CoT. Conversely, the theory suggests fine-tuning offers broader potential, theoretically enabling a sufficiently powerful model to solve any task within that category given adequate resources.

Diagram of Thought (DoT) builds upon these diverse approaches while offering key distinctions. Like ToT and GoT, DoT utilizes non-linear structures (DAGs) for reasoning. However, it distinctively internalizes the graph construction and navigation within a \emph{single} auto-regressive model via role tokens, minimizing external control dependencies. This contrasts with the external orchestration often required by ToT and GoT. DoT employs explicit cognitive roles (propose, critique, summarize), similar to CR, but integrates them seamlessly within one model through conditional generation, avoiding multi-agent coordination complexities. The use of rich natural language critiques potentially offers more nuanced feedback than the simple heuristic scores sometimes used in ToT. Importantly, by grounding the typed reasoning trace in Topos Theory, DoT aims for a level of formal rigor and consistency criteria that distinguishes it from purely heuristic methods, resonating with the structural analysis provided by works like \citet{yuan2023power}. DoT thus presents a unified, self-contained, interpretable, and formally grounded approach to advance complex reasoning in LLMs.

\section{The Diagram of Thought Framework}
\label{sec:dot_framework}

\begin{figure}[ht!]
\begin{center}
\centerline{\includegraphics[width=0.75\textwidth]{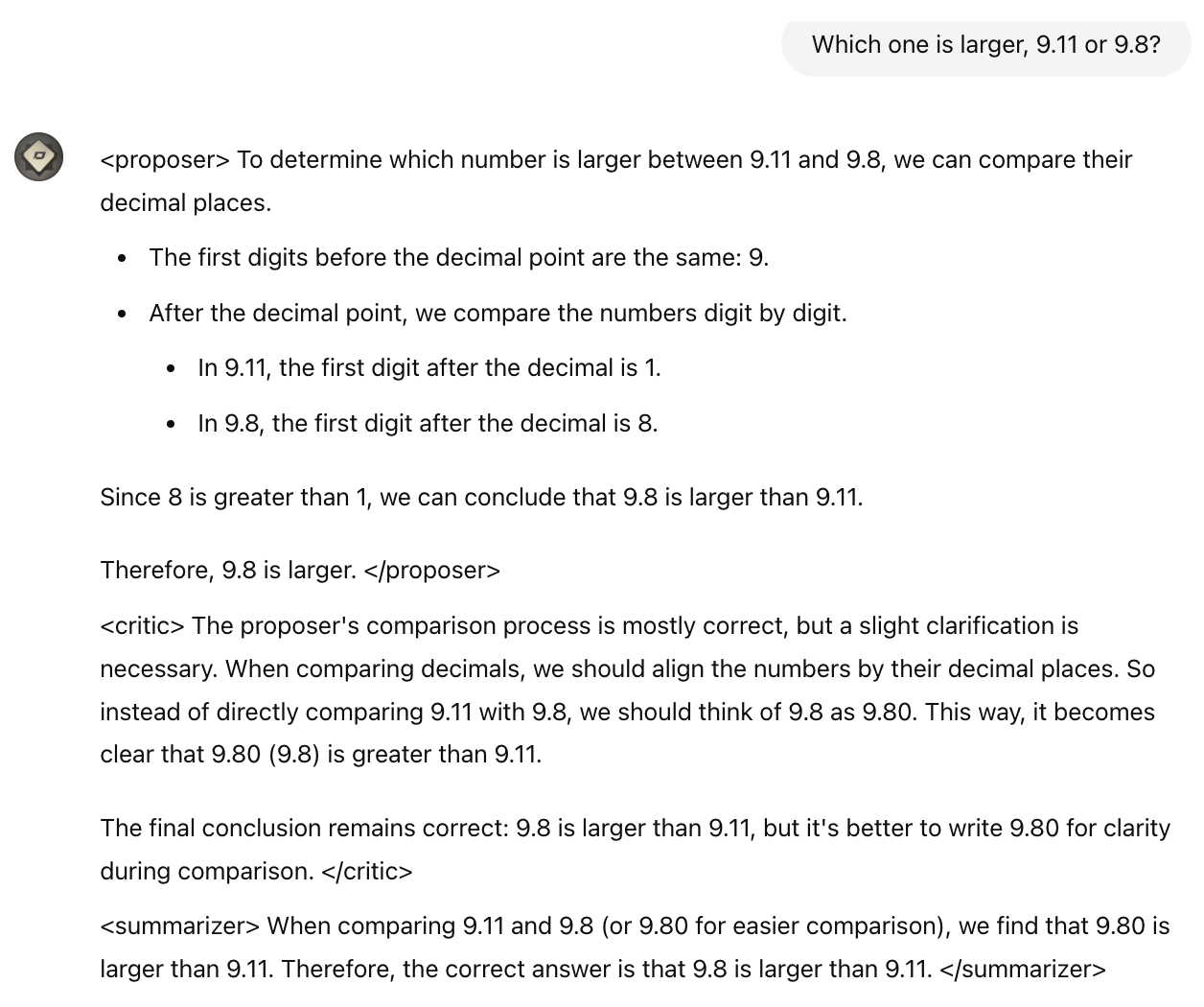}}
\caption{Illustrative example: Applying DoT reasoning steps to compare numerical values. Critiques might identify incorrect digit comparisons.}
\label{fig:numerical}
\end{center}
\end{figure}

\begin{figure}[ht!]
\begin{center}
\centerline{\includegraphics[width=0.75\textwidth]{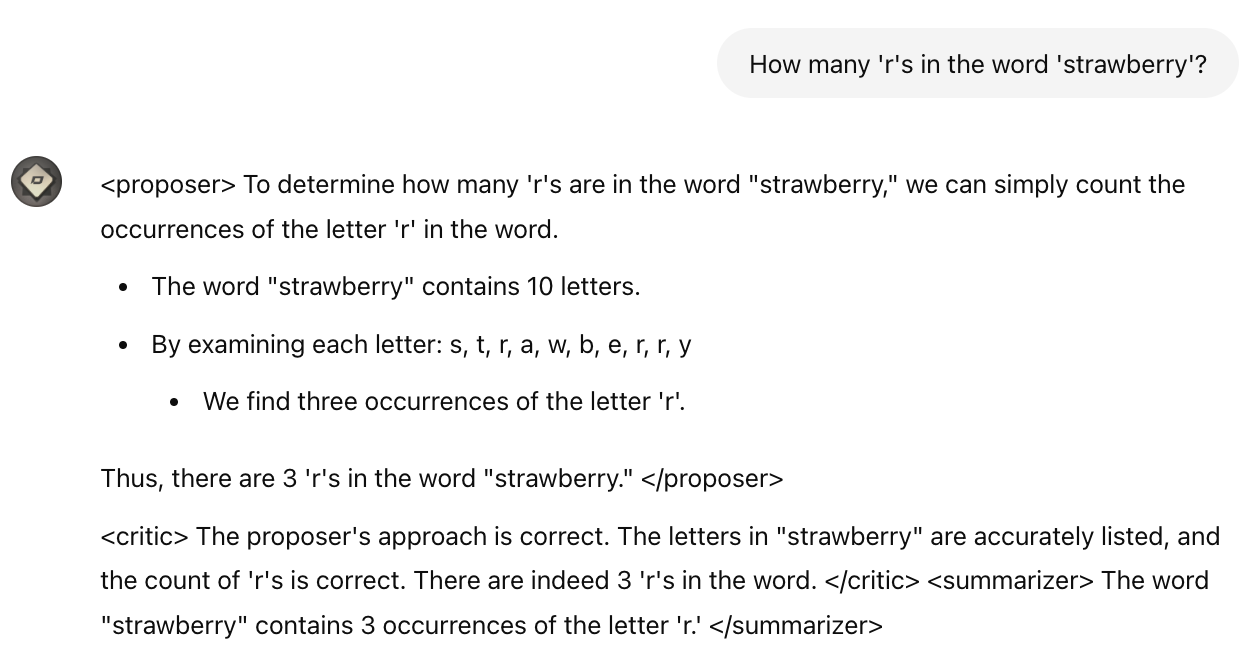}}
\caption{Illustrative example: A character-counting task where intermediate steps (identifying 'r's) and potential critiques (missed counts, double counts) could form a DoT graph.}
\label{fig:strawberry}
\end{center}
\end{figure}

In this section, we formally define the Diagram of Thought (DoT) framework. DoT operationalizes complex, iterative reasoning as the dynamic construction and traversal of a Directed Acyclic Graph (DAG) $G = (V, E)$ entirely within a single auto-regressive language model, $\Lang$. This internal graph structure allows the model to manage parallel lines of thought, critique intermediate steps, repair ideas based on feedback, and synthesize validated conclusions.

\begin{definition}[DoT Graph Components]
\label{def:dot_graph_components}
The DoT graph $G = (V, E)$ is composed of:
\begin{itemize}[parsep=1pt, itemsep=1pt, topsep=0pt,leftmargin=*]
\item \textbf{Nodes} $v \in V$: Each node represents a semantic unit or reasoning step. Every node $v$ is associated with:
\begin{itemize}
\item A specific role $\mathrm{role}(v) \in R = \{\text{Problem, Proposer, Critic, Summarizer}\}$.
\item Textual content $\mathrm{content}(v)$, generated by the LLM $\Lang$ while assuming the role $\mathrm{role}(v)$.
\item Optionally, an internal state $\mathrm{state}(v) \in \{\text{active, validated, invalidated, initial}\}$. For example, a `Proposer' node might start as `active', become `validated' after a positive critique, or `invalidated' after a negative critique.
\end{itemize}
\item \textbf{Edges} $(u, v) \in E$: A directed edge from node $u$ to node $v$ encodes that $v$ \emph{depends on} $u$. We standardize that, when emitting node $v$ (with fresh ID $v$), all edges in its block have \emph{dst}$=v$ and \emph{src}$<v$. This covers:
\begin{itemize}
\item Logical dependency (a new proposition depends on earlier premises). A node may depend on multiple sources, indicated by multiple edge records in its emission block.
\item Procedural dependency (a critic depends on the proposition it evaluates: proposer $\to$ critic).
\item Contextual dependency (a summarizer depends on the validated nodes it uses).
\end{itemize}
The operational provenance graph is acyclic by construction since every operational edge targets the current node and all sources are previously emitted nodes. This acyclicity statement concerns the provenance DAG only: the extracted semantic index category may contain arrows opposite to refinement provenance, isomorphisms, or cycles induced by certified equivalence records.
\end{itemize}
The construction of this graph is implicitly managed by the LLM's standard auto-regressive generation process, strategically guided by special role tokens and constrained by the validator whenever typed serialization is enabled.
\end{definition}

\subsection{Roles, Generation, and Iterative Reasoning}
\label{subsec:roles_autoregressive}

A core mechanism of DoT involves augmenting the LLM's vocabulary $\Vocab$ with a distinct set of role-specific tokens:
\[
T_{\text{roles}} = \{\texttt{<problem>}, \texttt{<proposer>}, \texttt{<critic>}, \texttt{<summarizer>}\}.
\]
Let $\Vocab' = \Vocab \cup T_{\text{roles}}$ be the augmented vocabulary. The LLM $\Lang$ operates by predicting the next token $w_t \in \Vocab'$ based on the preceding sequence (history) $H_{t-1} = w_1, \dots, w_{t-1}$:
\[
\Pr_{\Lang_\theta}(w_t \mid H_{t-1}) ,
\]
where $\theta$ denotes the model parameters. The generated sequence $H_T = w_1, \dots, w_T$ represents a serialized traversal and construction of the DoT graph $G$.

The role tokens function as control signals, prompting the LLM to adopt a specific cognitive function for the subsequent text generation, thereby determining the role and content of the next node(s) in the graph:

\begin{itemize}[parsep=1pt, itemsep=1pt, topsep=0pt,leftmargin=*]
\item \texttt{<problem>}: Typically precedes the initial problem statement $\mathcal{P}$. This establishes the root node $v_{\text{start}} \in V$ with $\mathrm{role}(v_{\text{start}}) = \text{Problem}$ and $\mathrm{content}(v_{\text{start}}) = \mathcal{P}$. Its state is $\mathrm{state}(v_{\text{start}}) = \text{initial}$.

\item \texttt{<proposer>}: Signals the LLM to generate a hypothesis, intermediate reasoning step, or potential solution fragment $P_i$. This creates a new node $v_{P_i}$ with $\mathrm{role}(v_{P_i}) = \text{Proposer}$ and $\mathrm{content}(v_{P_i}) = P_i$. All dependencies are declared explicitly via serialized \texttt{@edge} records defined in \Cref{subsec:serialization_extraction}. The new node typically starts with $\mathrm{state}(v_{P_i}) = \text{active}$.

\item \texttt{<critic>}: Instructs the LLM to evaluate one or more preceding `Proposer' nodes. The LLM generates a critique $C_j$, assessing validity, identifying flaws, or suggesting improvements. This creates a node $v_{C_j}$ with $\mathrm{role}(v_{C_j}) = \text{Critic}$ and explicit dependency edges from the propositions to the critic. We adopt a monotone validation discipline:
\begin{itemize}
\item If the critique validates a proposition, its state transitions to `validated'. This state is absorbing under the single-assignment discipline.
\item If the critique identifies flaws, its state transitions to `invalidated'. This renders the proposition ineligible for summarization. The reasoning path is expected to continue from the critique node to generate a repair or alternative proposition.
\end{itemize}

\item \texttt{<summarizer>}: Prompts the LLM to synthesize a final answer or consolidated conclusion. The model is trained to condition its summary generation on validated propositions explicitly selected by \texttt{@edge kind=use} records into the summarizer node, which are accessible through the serialized history $H_{t-1}$. It performs a conceptual aggregation respecting the declared dependencies and generates the summary text $S$. This creates a final node $v_S$ with $\mathrm{role}(v_S) = \text{Summarizer}$ and explicit \texttt{@edge} records from the validated propositions that contributed to the summary. A full-run summary may select all validated propositions by convention. Generation often terminates after this step.
\end{itemize}

The LLM learns to predict appropriate role token transitions based on the entire preceding history $H_{t-1}$, effectively learning to navigate and structure the reasoning process. For instance, after generating a proposition via \texttt{<proposer>}, the model learns it is often appropriate to predict \texttt{<critic>}. Following a critical critique (\texttt{<critic>} leading to state `invalidated'), the model might predict \texttt{<proposer>} again to generate a repair, or explore an alternative branch.

The DoT reasoning process unfolds as the LLM generates a serialized representation of the DAG $G$:
\begin{enumerate}[parsep=1pt, itemsep=1pt, topsep=0pt,leftmargin=*]
\item \textbf{Initialization}: The process begins with the problem statement, formatted using the typed serialization from \Cref{subsec:serialization_extraction} (e.g., \texttt{@node id=1 role=problem ...}). This defines the root node $v_1$.
\item \textbf{Proposal}: The LLM predicts \texttt{<proposer>} and generates the text for a proposition $P_2$, along with its node definition (\texttt{@node id=2 role=proposer}) and an edge from a prior node (\texttt{@edge src=1 dst=2 kind=use}). This adds node $v_2$ (state: active) to $G$.
\item \textbf{Critique}: The LLM predicts \texttt{<critic>}, generates a critique $C_3$ for $P_2$, and emits the corresponding node and edge records (\texttt{@node id=3 role=critic}, \texttt{@edge src=2 dst=3 kind=critique}). It also emits a status record (\texttt{@status target=2 mark=validated|invalidated}) that updates the state of $v_2$.
\item \textbf{Continuation (Branching/Repair/Exploration)}: Based on the history (including the state of $v_2$), the LLM predicts the next role token:
\begin{itemize}
\item If $v_2$ was validated: The LLM might predict \texttt{<proposer>} to generate a new proposition $P_4$ building upon $v_2$ (adding node $v_4$ and an \texttt{@edge} from $v_2$).
\item If $v_2$ was invalidated by $C_3$: The LLM might predict \texttt{<proposer>} to generate a repaired proposition $P_4$ that addresses the critique, with a procedural edge from $v_3$ (\texttt{@edge src=3 dst=4 kind=refine}). Such a repair edge records provenance; it induces a semantic entailment arrow only if accompanied by a typed certificate such as \texttt{@entails}. Alternatively, it might backtrack to generate an alternative proposition $P_5$ stemming from the root node $v_1$.
\end{itemize}
\item \textbf{Iteration}: Steps 2--4 repeat, progressively extending the operational DAG. The requirement that operational edge destinations must have higher IDs than their sources syntactically enforces acyclicity of provenance.
\item \textbf{Summarization}: Eventually, the LLM predicts \texttt{<summarizer>}. It is trained to generate a summary by synthesizing information from selected validated nodes, adding a summary node, and explicit \texttt{@edge} records from the nodes it used.
\end{enumerate}
This process yields a structured, interpretable trace of reasoning, captured within a single, self-contained generated sequence.

\subsection{Typed Serialization, Validation, and Extraction}
\label{subsec:serialization_extraction}

To ground the reasoning process and ensure auditability, we introduce a disciplined, typed serialization format. Natural language content is interleaved with structured records (prefixed with `@') that define the DAG's nodes, edges, and states. For example, \texttt{@node id=3 role=critic} creates a critic node, and \texttt{@edge src=2 dst=3 kind=critique} establishes its dependency on a previous proposition. Node IDs are strictly increasing, guaranteeing acyclicity of the operational provenance DAG by construction.

During inference, a lightweight, controller-light validator enforces this structure. It uses grammar- and register-constrained masking to ensure the LLM only generates syntactically valid records (e.g., an edge's source must be a previously defined node) and, where typed blocks are present, calls the selected solver checks for semantic certificates. This process is deterministic and avoids external search algorithms or planners. A deterministic extraction map, $\Phi$, converts any well-formed typed trace into a formal syntactic diagram, and then into a semantic diagram after a typed interpretation is fixed, enabling the categorical semantics described in Section \ref{sec:topos_formalization}. The full grammar, operational semantics, and validation rules are detailed in Appendix \ref{app:serialization}.

\subsection{Training and Controller-Light Inference}
\label{subsec:training_inference}

\textbf{Training}: The DoT capability is instilled in the LLM $\Lang$ through fine-tuning on datasets formatted according to the DoT structure. Such data consists of sequences $H = w_1, \dots, w_T$ containing appropriately interleaved role tokens ($T_{\text{roles}}$) and natural language text segments, representing valid, coherent reasoning DAGs. Potential data sources include:
\begin{itemize}[parsep=1pt, itemsep=1pt, topsep=0pt, leftmargin=*]
\item Curated examples derived from human step-by-step reasoning traces, augmented with role tokens and serialized records.
\item Synthetically generated examples from structured problem-solving processes (e.g., program execution traces, mathematical proofs), formatted using the typed serialization from \Cref{subsec:serialization_extraction}.
\item Bootstrapped data generated by an initial version of a DoT model, potentially filtered or refined based on correctness or coherence metrics.
\end{itemize}
The training objective is the standard auto-regressive language modeling loss (e.g., cross-entropy) applied over the entire sequence, including role tokens and content tokens:
\[
\mathcal{L}(\theta) = - \frac{1}{|H|}\sum_{t=1}^{|H|} \log \Pr_{\Lang_\theta}(w_t \mid w_1, \dots, w_{t-1}).
\]
This objective trains the model $\Lang$ to simultaneously learn the reasoning patterns associated with each role (proposing, critiquing, summarizing) and the appropriate transitions between these roles based on the context, thereby internalizing the ability to construct and navigate the DoT graph structure.

\noindent\textbf{Inference}: To solve a new problem $\mathcal{P}$ using DoT, inference proceeds as follows:
\begin{enumerate}[parsep=1pt, itemsep=1pt, topsep=0pt,leftmargin=*]
\item Initialize the generation history $H$ with the serialized problem statement, e.g., \texttt{@node id=1 role=problem ...}.
\item Perform auto-regressive generation using the trained LLM $\Lang$. At each step $t$, sample or select the next token $w_t \sim \Lang(H_{t-1})$ using a chosen decoding strategy (e.g., greedy decoding, nucleus sampling) and the well-formedness constraints from \Cref{subsec:serialization_extraction}.
\item Append the generated token $w_t$ to the history: $H_t = H_{t-1} \oplus w_t$.
\item Repeat steps 2 and 3 until a termination condition is met. Common conditions include:
\begin{itemize}
\item Generation of the \texttt{<summarizer>} token and its subsequent content, followed by a special end token.
\item Reaching a predefined maximum sequence length or generation budget.
\item Generation of a specific end-of-sequence token.
\end{itemize}
\end{enumerate}
The final output is typically the textual content associated with the \texttt{<summarizer>} node, although the complete generated sequence $H$ provides the full reasoning trace (the serialized DoT graph) for interpretability. Notably, this entire process is self-contained within the single LLM $\Lang$; no external graph management system or search/planning controller is required during inference, beyond a deterministic syntax validator and, when typed blocks are present, a decidable entailment check for the chosen typed fragment.

\section{Topos-Theoretic Formalization of DoT}
\label{sec:topos_formalization}

\paragraph{Order convention.}
We write $\Sub(S)$ for the Heyting algebra of subobjects of $S$, ordered by inclusion ($P\le Q$ iff $P\hookrightarrow Q$).
Under this order, more informative / more constrained propositions correspond to smaller subobjects.
This order-theoretic presentation applies to the predicate/mono instantiation in which proposer nodes are interpreted as monomorphisms into $S$.
To align “more information” with “larger” elements, we work in the opposite poset
\[
\Pred(S) := \Sub(S)^{\op},
\]
so that $P \preceq Q$ in $\Pred(S)$ iff $Q\le P$ in $\Sub(S)$.

For a finite discrete family, the colimit in $\Pred(S)$ is its join; translated back to $\Sub(S)$ this is a meet (intersection):
\[
\colim_{\Pred(S)} \{P_v\}_{v\in V}
\;=\;
\bigvee\nolimits_{\Pred(S)} \{P_v\}_{v\in V}
\;=\;
\bigwedge\nolimits_{v\in V} P_v
\quad\text{(computed in }\Sub(S)\text{)}.
\]
The empty meet is $S$, the terminal object of the slice $(\E/S)$.
For an actual semantic diagram $D:\J\to\Sub(S)$, the variance is explicit: the corresponding information-order diagram is $D^{\op}:\J^{\op}\to\Pred(S)=\Sub(S)^{\op}$, and
\[
\colim\nolimits_{\Pred(S)}D^{\op}\;\cong\;\lim\nolimits_{\Sub(S)}D.
\]
Equivalently (and this is the formulation we use in the general-arrow setting), conjunction-like synthesis is a finite limit in the slice $(\E/S)$: for subobjects, finite limits are pullbacks and compute intersections. We keep the posetal fragment explicit (Assumption~\ref{ass:mono}); the general-arrow case is handled via slice sheafification together with finite limits (Cor.~\ref{cor:general_sheafification}).

\smallskip
\noindent\textbf{General slice instantiation.} When proposer nodes are interpreted as general objects $X\to S$ in $(\E/S)$ (not necessarily monos), the poset $\Sub(S)$ no longer captures the whole semantics. Synthesis is then the finite limit of the extracted selected diagram in $(\E/S)$, not merely a meet of its objects; validation reflects this finite limit into the sheaf slice via the induced left-exact functor $a_j^{/S}:(\E/S)\to(\mathrm{Sh}_j(\E)/a_jS)$. The predicate/mono equations are recovered by restricting to monomorphisms and comparing the sheafified result back over $S$ by pullback along the unit $\eta_S:S\to a_jS$.

While the operational description in Section \ref{sec:dot_framework} details the DoT mechanism, establishing its logical soundness and robustness requires a deeper, formal framework. We leverage Topos Theory \citep{maclane2012sheaves, johnstone2002sketches, lambek1988introduction}, which provides a setting with finite limits, exponentials, and a subobject classifier to interpret intuitionistic logic. This makes it suitable for formalizing the dynamic, evidence-aggregating, and context-dependent reasoning inherent in the DoT process.

An elementary topos $\E$ is a category that encapsulates key properties needed for modeling logical systems and computation:
\begin{enumerate}[parsep=1pt, itemsep=1pt, topsep=0pt, leftmargin=*]
\item \textbf{Finite Limits:} $\E$ has a terminal object $1$, binary products $A \times B$, and pullbacks. This allows for combining and constraining information.
\item \textbf{Cartesian Closure:} For any objects $A, B \in \E$, there exists an exponential object $B^A$, representing the internal collection of morphisms $A \to B$. This enables modeling functions, predicates, and higher-order logic.
\item \textbf{Subobject Classifier $\Omega$:} There exists an object $\Omega$ and a truth arrow $\top: 1 \to \Omega$ such that every monomorphism $m: P \hookrightarrow A$ corresponds uniquely to a characteristic morphism $\chi_m: A \to \Omega$. $\Omega$ internalizes the logic of the topos, which is generally a Heyting algebra, supporting intuitionistic reasoning.
\end{enumerate}
Crucially for DoT, presheaf topoi $\E = \Set^{\C^{\mathrm{op}}}$ are complete and cocomplete. For the synthesis step we focus on, the relevant universal construction is a finite limit (conjunction of constraints) in the slice $(\E/S)$; dually, this is a variance-reversed colimit in the information order $\Pred(S)=\Sub(S)^{\op}$. Since slices of a presheaf topos are again topoi, the slice $(\E/S)$ has the finite limits we require (and also all small colimits, should one wish to model disjunctive/quotient-style aggregation).

In what follows, we \emph{fix} a presheaf topos $\E = \Set^{\C^{\mathrm{op}}}$. This is the category of functors $\C^{\op}\to\Set$ for a small category $\C$. We also fix a designated semantic object $S \in \E$ representing the universe of discourse. Our formalization takes place within the \textit{slice category} $(\E/S)$, where predicate-like propositions are subobjects of $S$ and general typed propositions are objects over $S$. For concreteness, one may take $\C$ to index problem contexts and $S(c)$ to be the set of admissible semantic states at context $c$. For the QF-LIA instantiation used in examples, we take $\C$ to be a finite discrete category (a single object in simple cases) and interpret terms componentwise so that interpretation remains decidable.

\begin{definition}[Categorical Semantics of DoT Components]
\label{def:dot_components_topos}
Within the fixed presheaf topos $\E=\Set^{\C^{\mathrm{op}}}$ and slice $(\E/S)$, fix a Lawvere--Tierney topology $j:\Omega\to\Omega$ with associated universal closure operator $c=(c_A)_{A\in\E}$ on subobjects (extensive, idempotent, monotone, pullback-stable). When discussing only subobjects of $S$, we write $c$ for the component $c_S:\Sub(S)\to\Sub(S)$.
\begin{itemize}[parsep=1pt, itemsep=1pt, topsep=0pt,leftmargin=*]
\item \textbf{Semantic Space ($S$):} A base object $S \in \E$ representing the universe of discourse or the space of all possible solutions and intermediate states.
\item \textbf{Propositions ($P$):} A proposer node is interpreted as an object $p:P\to S$ of the slice category $(\E/S)$. In the predicate/mono instantiation, $p$ is a monomorphism $P\hookrightarrow S$, and we freely identify it with the corresponding subobject (a ``subset'' of $S$).
\item \textbf{Entailment vs.\ dependency variance:} In the predicate instantiation, entailment $P \Rightarrow Q$ corresponds to inclusion $P\le Q$ in $\Sub(S)$ (equivalently, the unique morphism $P\to Q$ over $S$). Operational dependency edges are not, by themselves, semantic entailments. The extracted index category $\J_G$ (Def.~\ref{def:index_cat}) generates semantic arrows only from certified strengthening/refinement records and certified \texttt{@entails} records. Thus a certified strengthening of proposer $u$ by proposer $v$ induces an arrow $v\to u$ in $\J_G$, interpreted as a map $D_G(v)\to D_G(u)$ over $S$ (predicate case: $D_G(v)\le D_G(u)$), while \texttt{@entails src=i dst=k} induces an arrow $i\to k$ interpreted as $D_G(i)\to D_G(k)$. In particular, an operational record \texttt{@edge src=u dst=v kind=refine} points from the old node to the new node as provenance, but if it carries a typed strengthening certificate, the induced semantic arrow is $v\to u$. Ordinary \texttt{use}, \texttt{critique}, and uncertified repair-style \texttt{refine} edges record provenance/control flow only.
\item \textbf{Critiques as judgements (typed):} A \texttt{<critic>} node is typed evidence \emph{about} one or more propositions. Semantically, an accepted critique record contributes (i) a validation mark for its target proposer, and optionally (ii) typed arrows, isomorphisms, or path-equality constraints between proposer objects over $S$ (e.g., certified \texttt{@entails}/\texttt{@eq} records), which become generators or relations in the extracted diagram. In the predicate/mono fragment, certified entailment arrows are inclusions between subobjects; in the general slice fragment, they may be arbitrary morphisms over $S$. Critic nodes themselves are witnesses for these records rather than objects of the semantic synthesis diagram.
\item \textbf{Validation as a nucleus/reflection:} On subobjects, validation is modeled by the universal closure $c:\mathrm{Sub}(S)\to\mathrm{Sub}(S)$ induced by $j$; in particular $c$ is extensive, monotone, idempotent, pullback-stable, and satisfies $c(P\wedge Q)=c(P)\wedge c(Q)$ for finite meets (i.e., $c$ is a nucleus). In the predicate/mono fragment, a proposition is semantically validated iff $c(P)=P$ (i.e., it is $c$-closed). For a general slice object $X\to S$, validation is represented by applying the induced left-exact sheafification functor $a^{/S}_j$; we do not call arbitrary objects $c$-closed unless they are subobjects.
\item \textbf{Strengthening vs.\ repair:} A strengthening step produces a new proposer object $p':P'\to S$ equipped with a certified morphism $P'\to P$ over $S$ (interpreted as ``$P'$ entails/strengthens $P$''). In the predicate/mono fragment, this is precisely an inclusion $P'\hookrightarrow P$ in $\Sub(S)$. A repair step after an invalidation need not strengthen the rejected proposition; it contributes to the semantic diagram only through separately certified entailment or equivalence records.
\item \textbf{Selected Validated Diagram:} Only validated proposer nodes selected by the summarizer enter the synthesis computation; critique nodes provide certified morphisms and equivalence constraints between proposer nodes.
\end{itemize}
\end{definition}

\begin{assumption}[Fixed, pullback-stable validation modality]
\label{ass:j-topology}
There exists a Lawvere--Tierney topology $j:\Omega\to\Omega$ on $\E$ whose induced universal closure operator $c=(c_A)_{A\in\E}$ models validation. Each component $c_A:\Sub(A)\to\Sub(A)$ is extensive, monotone, idempotent, and pullback-stable. All results in this section are relative to this fixed $j$.
Moreover, since $c$ is induced by a Lawvere--Tierney topology, each $c_A$ is a nucleus on $\Sub(A)$: it preserves finite meets. When we apply $c$ to propositions over $S$, we mean the component $c_S$.
We work throughout in the presheaf topos setting, so the finite limits we use in synthesis exist in $(\E/S)$; the induced functor $a^{/S}_j$ is left exact and therefore preserves these finite limits.
\end{assumption}

\subsection{Semantic Target and Normative Conditions}
The following assumptions connect the practical LLM behavior with our formal model. They are idealized, normative conditions that define the target semantics for a well-behaved DoT agent. The operational mechanisms are designed to make LLM traces more likely to satisfy these conditions upon interpretation.
\begin{enumerate}[parsep=1pt, itemsep=1pt, topsep=1pt,label=(A\arabic*)]
\item \label{ass:interp} Typed proposer nodes admit interpretations as objects over $S$ (as subobjects in the predicate/mono fragment); certified critique content interprets as arrows/predicates that constrain these objects. Certified strengthening/refinement and \texttt{@entails} records correspond to morphisms over $S$.\footnote{This is a strong assumption; our framework is normative: it specifies the target semantics an ideal DoT agent should realize, enabled by the typed serialization in \Cref{subsec:serialization_extraction}.}
\item \label{ass:closure} In the predicate/mono fragment, critique-driven validation corresponds to the Lawvere--Tierney closure $c$ on $\mathrm{Sub}(S)$; validated predicate nodes are exactly the $c$-closed subobjects. In the general-arrow fragment, validated synthesis is computed after applying the induced slice sheafification functor.
\item \label{ass:coh} Equivalences and path equalities identified by validated critiques are respected. Formally, we quotient the generated index category by the smallest congruence induced by certified equality records (see Def.~\ref{def:index_cat}); in the predicate/posetal syntax, \texttt{@eq src=i dst=k} abbreviates mutual entailment/equality of proposer subobjects, while in the general slice syntax it requires a certified isomorphism over $S$.
\item \label{ass:mono} For our main result, we consider the fragment in which every selected validated proposer is interpreted as a monomorphism $P\hookrightarrow S$ and every selected validated arrow between proposers is (hence) an inclusion in $\Sub(S)$, so the selected diagram lands in the posetal category $\Sub(S)\subseteq(\E/S)$.
\end{enumerate}

\subsection{The Reasoning Diagram and its Synthesis via Finite Limit}

The DoT-DAG $G=(V, E)$ specifies the operational trace. The categorical diagram used for synthesis is extracted from the certified semantic records inside that trace.

\begin{definition}[DoT Index Category $\J_G$]
\label{def:index_cat}
Given a well-formed typed trace with DoT DAG $G = (V, E)$, let $V_{\mathrm{prop}}\subseteq V$ be the subset of proposer nodes. We distinguish the operational DAG from the semantic diagram: ordinary dependency edges do not automatically generate semantic morphisms.

Let $\mathsf A_{\mathrm{sem}}(G)$ be the directed multigraph on $V_{\mathrm{prop}}$ with the following certified semantic generators:
\begin{itemize}[parsep=1pt, itemsep=1pt, topsep=0pt,leftmargin=*]
\item an arrow $v\to u$ whenever the trace contains an accepted typed certificate that proposer $v$ strengthens proposer $u$; for a certified operational record \texttt{@edge src=u dst=v kind=refine}, the semantic arrow is therefore opposite to the operational dependency edge;
\item an arrow $i\to k$ for each accepted record \texttt{@entails src=i dst=k} between proposer nodes;
\item in the general slice fragment, inverse arrows for each accepted \texttt{@eq} record certified as an isomorphism over $S$.
\end{itemize}
Pure \texttt{use}, \texttt{critique}, and uncertified repair edges are excluded from $\mathsf A_{\mathrm{sem}}(G)$.

The semantic index category is
\[
\J_G := \mathsf{Free}(\mathsf A_{\mathrm{sem}}(G))/\!\equiv,
\]
where $\equiv$ is the smallest arrow congruence generated by certified path-equality records and, for certified isomorphism records, by the inverse equations. In the predicate/posetal fragment, \texttt{@eq src=i dst=k} abbreviates mutual entailment and hence equality of subobjects; equivalently, it identifies the corresponding objects in the thin category. Richer path-equality syntax can be added by extending the same congruence.

For all validated content, $\J_{\mathrm{valid}}$ denotes the finite presentation generated by validated proposer objects and certified semantic arrows whose source and target are validated. Paths through invalidated proposer nodes are not used unless the trace separately certifies an arrow between validated endpoints.

\smallskip
For a particular summarizer node $s$, let
\[
V_{\Sigma}(s)=\{\,v\in V_{\mathrm{prop}}:\ v\text{ is validated and } \texttt{@edge src=}v\ \texttt{dst=}s\ \texttt{kind=use}\text{ is accepted}\,\}.
\]
The synthesis presentation $\J_{\Sigma}(s)$ is the subpresentation generated by $V_{\Sigma}(s)$ and the certified semantic arrows and relations whose endpoints lie in $V_{\Sigma}(s)$. If no summarizer node is specified and one wants a global full-run summary, we set $V_{\Sigma}=V_{\mathrm{valid}}$ and $\J_{\Sigma}=\J_{\mathrm{valid}}$ by convention.
\end{definition}

\begin{theorem}[DoT Process as Diagram Construction]
\label{thm:diagram_construction}
A DoT reasoning process generating a well-formed typed DAG $G=(V,E)$, together with a fixed semantic interpretation of typed propositions and certified arrows, defines a functor (a diagram) $D_G: \J_G \to (\E/S)$ with:
\begin{itemize}[parsep=1pt, itemsep=1pt, topsep=0pt, leftmargin=*]
\item Each typed proposer node $v$ mapped to an object $D_G(v)\to S$ in the slice (a subobject $D_G(v)\hookrightarrow S$ in the predicate/mono fragment);
\item Accepted critic records supply certified arrows, isomorphisms, or path equalities between proposer objects; critic nodes are witnesses for these records rather than objects of $D_G$;
\item Each arrow $v\to u$ in $\J_G$ is mapped to a certified morphism over $S$, witnessing entailment or strengthening coherence (posetal case: an inclusion $D_G(v)\hookrightarrow D_G(u)$).
\end{itemize}
Functoriality ensures that composite certified dependencies (via paths modulo $\equiv$) are respected in the slice.
\end{theorem}
\begin{proof}[Proof Sketch]
By Assumption~\ref{ass:interp}, each typed proposer node $v$ admits an interpretation as an object $D_G(v)\to S$ of the slice $(\E/S)$, and as a subobject in the predicate/mono fragment. The generators of $\J_G$ are certified strengthening/refinement arrows, certified \texttt{@entails} arrows, and certified isomorphism/path-equality records. The validator requires each such generator to have a typed semantic witness, so Assumption~\ref{ass:interp} assigns it a morphism over $S$. We define $D_G$ on composite arrows by composition in $(\E/S)$. Certified equality records generate the congruence used in $\J_G$, and Assumption~\ref{ass:coh} ensures that equivalent syntactic paths induce equal semantic composites. Hence $D_G$ is well-defined and preserves identities and composition.
\end{proof}

\noindent\textbf{Synthesis by information-colimit (slice-limit) and reflection.}
The \texttt{<summarizer>} aggregates selected validated content. We distinguish an inclusion/posetal setting (our main focus) and a general-arrow setting. In the latter, sheafification induces a base-change: $a^{/S}_j:(\E/S)\to(\mathrm{Sh}_j(\E)/a_j S)$, and summaries are read over $a_j S$ via the unit $S\to a_j S$.

\begin{theorem}[Summarization as finite meet-plus-closure in the reflective subposet]
\label{thm:summarization_colim_info}
Assume every proposer selected for synthesis is interpreted as a $c$-closed subobject of $S$ and every certified semantic arrow between selected proposers is an inclusion, so the selected diagram $D_{\Sigma}=D_G|_{\J_{\Sigma}}$ lands in the thin category $\Sub(S)$. Let $V_{\Sigma}$ be the finite set of validated proposer nodes selected by the summarizer, with the convention that the empty meet is $S$. The finite-limit summary is
\[
\mathrm{Summary}_{\Sigma}
\;=\;
\lim\nolimits_{\Sub(S)}D_{\Sigma}
\;=\;
\bigwedge_{v\in V_{\Sigma}}D_G(v).
\]
Equivalently, this is the colimit of the variance-reversed diagram
$D_{\Sigma}^{\op}:\J_{\Sigma}^{\op}\to\Pred(S)=\Sub(S)^{\op}$. If one writes the reflected form
\[
c\Big(\bigwedge_{v\in V_{\Sigma}}D_G(v)\Big),
\]
then the outer $c$ is redundant because $c$ is a nucleus and all inputs are already $c$-closed.
\end{theorem}
\begin{proof}[Proof Sketch]
In the predicate fragment, selected validated proposer nodes are subobjects of $S$. Since $\Sub(S)$ is thin, the limit of a finite diagram is the greatest lower bound of its objects, namely their meet. Equivalently, this meet is the finite join of the corresponding opposite diagram in the information order $\Pred(S)=\Sub(S)^{\op}$. Since each selected proposition is $c$-closed and $c$ preserves finite meets, the meet is again $c$-closed; writing the result as $c(\bigwedge_vD_G(v))$ emphasizes compatibility with the reflected construction.
\end{proof}

\begin{corollary}[General case via slice sheafification]
\label{cor:general_sheafification}
For a general selected diagram $D_{\Sigma}:\J_{\Sigma}\to(\E/S)$ (not necessarily posetal), presented by a finite generating graph and a finite set of certified coherence/path-equality relations already satisfied by $D_{\Sigma}$, the synthesized summary in the validated/sheaf slice is
\[
\mathrm{Summary}_{\Sigma}
\;\cong\;
\lim\nolimits_{\mathrm{Sh}_j(\E)/a_jS}\big(a^{/S}_j\circ D_{\Sigma}\big)
\;\cong\;
a^{/S}_j\!\big(\lim\nolimits_{\E/S} D_{\Sigma}\big),
\]
where $a^{/S}_j:(\E/S)\to(\mathrm{Sh}_j(\E)/a_j S)$ is the left-exact functor induced by sheafification, sending $X\to S$ to $a_jX\to a_jS$. If one wishes to compare the resulting object back to $(\E/S)$, one pulls back along the unit $\eta_S:S\to a_j S$. For a subobject $P\hookrightarrow S$, this inverse image represents the $j$-closure $c_S(P)\hookrightarrow S$; hence, in the posetal fragment, the pulled-back sheafified limit reduces to the reflected form in Theorem~\ref{thm:summarization_colim_info}.
\end{corollary}
\begin{proof}[Proof Sketch]
Since runs are finite, the extracted semantic shape has a finite graph presentation and finitely many certified path-equality relations. The accepted relations are part of the well-definedness of $D_{\Sigma}$; once they are satisfied, the limit can be computed from finite products and equalizers imposing cone compatibility for the finitely many generating arrows. The slice sheafification functor $a^{/S}_j$ is left exact (as $a_j$ is), hence preserves finite limits. In the posetal fragment, limits in $\Sub(S)$ are meets (intersections). The sheafified mono lives over $a_jS$; pulling it back along $\eta_S:S\to a_jS$ gives the $j$-closure of the original meet. Thus the subobject seen back over $S$ is $c_S(\bigwedge_vD_G(v))$, recovering Theorem~\ref{thm:summarization_colim_info}.
\end{proof}

\subsection{Formal Guarantees: Consistency and Robustness}

\begin{theorem}[Conditional consistency via closure validation]
\label{thm:consistency_guarantee}
Fix $\E=\Set^{\C^{\mathrm{op}}}$ and the Lawvere--Tierney closure $c$ on $\Sub(S)$. For a finite DoT run and a chosen summarizer, let $V_{\Sigma}$ be the finite set of validated, $c$-closed subobjects selected for synthesis. The following two consistency readings are valid:
\begin{enumerate}[parsep=1pt, itemsep=1pt, topsep=0pt, leftmargin=*]
\item \textbf{Fixed-stage finite family:} If the interpreted family is jointly satisfiable relative to a fixed stage $a\in\C$, i.e. there exists $x\in S(a)$ lying in every $D_G(v)(a)$, then
\[
c\!\left(\bigwedge_{v\in V_{\Sigma}} D_G(v)\right)(a)\neq\varnothing
\]
and the summary is non-initial. Componentwise non-emptiness of the individual $D_G(v)(a)$ is not sufficient; the premise requires a common element.
\item \textbf{Model-compact setting:} Let $\Gamma$ be the set of first-order formulas represented by an idealized validated typed family, and suppose $\mathbb{T}$ is a first-order background theory. If every finite subset of $\Gamma$ is satisfiable together with $\mathbb{T}$, then compactness gives a model $\mathfrak M\models\mathbb{T}$ and an assignment satisfying all formulas in $\Gamma$. Thus the conjunction represented by the family is satisfiable in that model. If the chosen semantic interpretation of $S$ and $c$ is evaluated in the same model, extensivity of $c$ preserves satisfiability of the corresponding closed summary. This is a model-theoretic satisfiability statement, not a claim of satisfiability in the intended/standard model, and not a claim of componentwise non-emptiness in a fixed presheaf stage.
\end{enumerate}
\end{theorem}
\begin{proof}[Proof Sketch]
Case (1): at the fixed stage $a$, joint satisfiability gives a common element of all interpreted subobjects, so the componentwise finite intersection is non-empty; extensivity of $c$ preserves non-emptiness. Case (2) is the standard compactness theorem for first-order logic, followed by the observation that an extensive closure cannot destroy a realizing assignment when the closure is interpreted in the same semantic structure.
\end{proof}

\begin{corollary}[Soundness w.r.t. satisfiability]
If $[\![\cdot]\!]$ maps into $\Sub(S)$ and the selected validated family $\{P_v\}_{v \in V_{\Sigma}}$ is jointly satisfiable in the same fixed stage or semantic structure in which the summary is interpreted, then the meet $\bigwedge_v P_v$ is satisfiable. Consequently, by extensivity, the reflected summary $c\left(\bigwedge_v P_v\right)$ is satisfiable in that same interpretation.
\end{corollary}

\begin{remark}[Internal Logic, Modality, and Variance]
The internal logic of $\E$ equips $\mathrm{Sub}(S)$ with a Heyting algebra. Validation via a Lawvere--Tierney topology promotes propositions to $c$-closed ones. In inclusion order, synthesis is a finite meet of constraints; after reversing variance, the corresponding diagram in the opposite information order $\Pred(S)=\Sub(S)^{\op}$ has this same object as a finite colimit. Thus ``gluing validated parts'' should be read as a finite slice-limit, or equivalently as an information-order colimit in the predicate fragment, not as an ordinary colimit in $\Sub(S)$.
\end{remark}

\begin{proposition}[Robustness under Diagram Isomorphisms]
\label{prop:robustness_universality}
Let $D_{\Sigma,1}: \J_{\Sigma,1} \to (\E/S)$ and $D_{\Sigma,2}: \J_{\Sigma,2} \to (\E/S)$ represent selected validated reasoning steps from two different runs. If there is an isomorphism of diagrams, then their slice-limits (equivalently, information-colimits in the predicate fragment) are isomorphic:
\[
\lim D_{\Sigma,1} \cong \lim D_{\Sigma,2}.
\]
This implies that the synthesized semantic content depends on the abstract structure of the selected validated reasoning diagram, not on incidental variations that preserve this structure.
\end{proposition}
\begin{proof}[Proof Sketch]
This is a direct consequence of the universal property of a limit. If two diagrams are isomorphic, there is a canonical isomorphism between their respective limits.
\end{proof}

\subsection{Immediate Consequences}
\label{subsec:immediate_consequences}

The formalization of the DoT synthesis step as a colimit in the variance-reversed information order (equivalently, as meet-plus-closure under inclusion) entails several immediate and desirable properties, grounded in the lattice-theoretic structure of subobjects and the properties of the closure operator $c$.

\begin{proposition}[Properties of Synthesis]
\label{prop:synthesis_properties}
Let $\mathcal{P} = \{P_v\}_{v\in V_{\Sigma}}$ be a finite selected family of predicate-fragment subobjects. Define the reflected synthesis operation by $\mathrm{Summary}(\mathcal{P}) = c(\bigwedge_{P \in \mathcal{P}} P)$; when all selected inputs are validated, this equals the ordinary meet because they are $c$-closed and $c$ is a nucleus. The operation exhibits the following properties:

\begin{enumerate}[itemsep=2pt, topsep=3pt, parsep=1pt,leftmargin=*]
\item \textbf{Finiteness in practice:} as runs are finite, all meets are finite; infinitary variants require compactness assumptions and are outside our core claims.

\item \textbf{Monotonicity (information order):} Adding a new selected proposition $P_{\text{new}}$ can only refine (never weaken) the reflected summary. In inclusion order,
\[
\mathrm{Summary}(\mathcal{P} \cup \{P_{\text{new}}\}) \subseteq \mathrm{Summary}(\mathcal{P}).
\]

\item \textbf{Idempotence (finite meets):} The system is robust to redundant validation. Re-processing already validated information over finite families does not alter the conclusion:
\[
\mathrm{Summary}(\mathcal{P}) = c\left( \bigwedge_{P \in \mathcal{P}} c(P) \right) = c\left( \bigwedge_{P \in \mathcal{P}} P \right).
\]
The first equality uses finite-meet preservation of $c$; when validated propositions are already $c$-closed ($P = c(P)$), the statement is immediate.

\item \textbf{Conservativity (Redundancy Elimination):} If a validated proposition $P_w$ is already \emph{no stronger than} the others’ conjunction (i.e., $P_w \supseteq \bigwedge_{P \in \mathcal{P}\setminus\{P_w\}} P$), its explicit inclusion does not change the summary.
\[
\mathrm{Summary}(\mathcal{P}) = \mathrm{Summary}(\mathcal{P} \setminus \{P_w\}).
\]
\end{enumerate}
\end{proposition}
\begin{proof}
These properties are direct consequences of the underlying mathematics. Monotonicity follows from the monotonicity of $\bigwedge$ and $c$. Idempotence follows from finite-meet preservation of $c$, and is immediate for already $c$-closed inputs. Conservativity follows because if $P_w$ contains the meet of the others, it does not change that meet.
\end{proof}

\begin{proposition}[Greatest $c$-closed Lower Bound (Canonicity)]
\label{prop:least_closed}
Let $\mathcal P=\{P_i\}_{i=1}^n\subseteq \Sub(S)$ be a finite family of selected validated subobjects (so $c(P_i)=P_i$ for all $i$),
and assume $c$ is a nucleus (finite-meet preserving; Assumption~\ref{ass:j-topology}).
Then the meet $\bigwedge_{i=1}^n P_i$ is itself $c$-closed and is the \emph{greatest $c$-closed lower bound} of $\mathcal P$
(in inclusion order). In particular,
\[
c\Big(\bigwedge_{i=1}^n P_i\Big)=\bigwedge_{i=1}^n P_i.
\]
\end{proposition}
\begin{proof}
Since $c$ preserves finite meets and fixes each $P_i$, we have
\[
c(\bigwedge_i P_i)=\bigwedge_i c(P_i)=\bigwedge_i P_i,
\]
so the meet is $c$-closed.
If $X$ is $c$-closed and $X\le P_i$ for all $i$, then $X\le \bigwedge_i P_i$ by the universal property of the meet.
\end{proof}

\begin{proposition}[Generalization of Linear Reasoning]
\label{prop:cot_special_case}
A linear Chain-of-Thought (CoT) process corresponds to a special case of a DoT diagram. Suppose the selected, operationally ordered validated propositions are $P_1,\dots,P_n$, and each later step strengthens the previous one, so that
\[
P_1 \supseteq P_2 \supseteq \dots \supseteq P_n
\quad\text{in }\Sub(S).
\]
Equivalently, the semantic inclusion arrows point from the later stronger proposition to the earlier weaker one:
\[
P_n \longrightarrow P_{n-1}\longrightarrow \cdots \longrightarrow P_1.
\]
Then the synthesis simplifies to the final step:
\[
\mathrm{Summary}(\{P_1, \dots, P_n\}) = P_n.
\]
\end{proposition}
\begin{proof}
For a chain $P_1 \ge \dots \ge P_n$ under inclusion, their meet is $\bigwedge_{i=1}^n P_i = P_n$. Applying $c$ gives $\mathrm{Summary} = c(P_n)$. Since $P_n$ is validated, it is $c$-closed, so $c(P_n)=P_n$.
\end{proof}

\begin{proposition}[Composition of Branches]
\label{prop:independent_branches}
For any two finite selected families $A$ and $B$, the overall summary of their union is the validated conjunction of their individual summaries (equivalently, the join in the information order). This identity is lattice-theoretic and does not require probabilistic or causal independence.
\[
\mathrm{Summary}(A \cup B)
= c(\mathrm{Summary}(A) \wedge \mathrm{Summary}(B))
= \mathrm{Summary}(A)\wedge \mathrm{Summary}(B).
\]
\end{proposition}
\begin{proof}
By definition, $\mathrm{Summary}(A \cup B) = c\!\left(\left(\bigwedge_{v \in A} P_v\right) \wedge \left(\bigwedge_{w \in B} P_w\right)\right)$. Since $c$ is a nucleus (finite-meet preserving), $c(X\wedge Y)=c(X)\wedge c(Y)$, and thus
\[
\mathrm{Summary}(A \cup B) \;=\; c\!\left(\bigwedge_{v \in A} P_v\right)\ \wedge\ c\!\left(\bigwedge_{w \in B} P_w\right)
\;=\;\mathrm{Summary}(A)\wedge \mathrm{Summary}(B),
\]
which is equivalent to the displayed formula because both individual summaries are $c$-closed.
\end{proof}

\subsection{Bridging Formalism and LLM Generation}

It is crucial to understand the relationship between this formal topos-theoretic model and the actual behavior of an LLM. The LLM does not explicitly perform computations within a topos. Instead:
\begin{itemize}[parsep=1pt, itemsep=1pt, topsep=1pt, leftmargin=*]
\item The topos framework provides the normative semantic model. It defines what constitutes sound, consistent, and robust synthesis. Theorems \ref{thm:summarization_colim_info}, \ref{thm:consistency_guarantee}, and Proposition \ref{prop:robustness_universality} describe desirable properties of an ideal reasoning process.
\item The LLM, trained on DoT-structured data (using the serialization from \Cref{subsec:serialization_extraction}), learns to generate text sequences that functionally approximate the operations described by the formalism. The generated sequence induces an abstract diagram that is then interpreted under Assumptions \ref{ass:interp}--\ref{ass:mono}.
\item Specifically, the \texttt{<summarizer>} role learns to generate text that effectively acts like the finite-limit/information-colimit described above: it synthesizes information from selected validated precursor nodes, respects their certified dependencies, and aims for a coherent, non-redundant aggregation.
\item The fidelity of this approximation depends heavily on the training data and model capacity. The topos model provides a precise target against which the LLM's reasoning behavior can be evaluated. One could design specific training objectives, such as a discriminative loss that penalizes generated summaries whose typed content violates the entailments dictated by the finite-limit construction.
\end{itemize}
This formalism offers a rigorous language for defining correctness criteria and provides a theoretical target for DoT behavior. Operational invalidation can either be modeled (i) as the absence of a validated inclusion (posetal fragment), or (ii) via a separate counterevidence object $I\in\Sub(S)$ with summaries computed as $c\big((\bigwedge \text{valid})\wedge \neg I\big)$ in the Heyting algebra $\Sub(S)$. We adopt the posetal refinement fragment for the main development and leave full revision semantics to future work.

\subsection{Separation from Linear Chain-of-Thought}
\label{subsec:separation}
\begin{theorem}[Structural separation from linear CoT]
\label{thm:separation}
Let $\E=\Set^{\C^{\mathrm{op}}}$, fix $S\in \E$, and assume validated arrows are inclusions in $\Sub(S)$ (posetal case). Suppose a DoT summarizer selects two validated, incomparable propositions $P,Q\in\Sub(S)$ (i.e., $P\not\le Q$ and $Q\not\le P$). Consider any attempt to faithfully embed this selected validated diagram into a linear Chain-of-Thought (a single chain of inclusions) via an order-preserving and order-reflecting functor that maps $P$ and $Q$ to distinct selected chain objects. No such embedding exists. Consequently, the two-branch DoT diagram cannot be represented as a faithful chain without changing the diagram; a linear trace may compute an equivalent conjunction only by adding or replacing nodes, not by embedding the original branching structure.
\end{theorem}
\begin{proof}[Proof sketch]
Interpret the selected validated proposer subdiagram as a poset-category. A linear CoT is a chain (total order) in $\Sub(S)$. To rule out degenerate collapse maps, we require an order embedding, i.e., a functor that is order-preserving and order-reflecting (equivalently, full and faithful for posets, and injective on the selected objects). In a chain, any two distinct images are comparable; by order-reflection this would force $P$ and $Q$ to be comparable in $\Sub(S)$, contradicting incomparability. Hence no order embedding of the selected validated DoT poset into a chain exists. The DoT summary is $c(P\wedge Q)$ by Thm.~\ref{thm:summarization_colim_info}; a chain can contain a later node denoting this meet, but doing so adds/replaces semantic content rather than faithfully embedding the original two-branch diagram.
\end{proof}

\section{Conclusion}
\label{sec:conclusion}

This paper introduced the Diagram of Thought (DoT), a framework that internalizes complex reasoning as DAG construction within a single auto-regressive LLM, guided by role-specific tokens and enforced by a lightweight, controller-light validator. We showed how DoT unifies proposition generation, critique, repair, and summarization without a heavyweight search/planning controller. We established a normative formalization using Topos Theory, where the synthesis of selected validated evidence corresponds to computing a finite limit in the slice (equivalently, a variance-reversed information-order colimit in the predicate fragment) and reflecting it along a Lawvere--Tierney topology when necessary.

Crucially, we moved beyond informal assumptions by specifying a typed serialization with online validation, decidable checks for selected typed fragments, a monotone state-update discipline, and support for multi-premise critiques. We separated operational critique records from the Lawvere--Tierney modality they are interpreted against. Our correctness claims are conditional on these checkable and auditable mechanisms, providing a solid bridge between the operational system and its semantic model.

This topos-theoretic perspective provides several key benefits:
\begin{itemize}[parsep=1pt, itemsep=1pt, topsep=0pt, leftmargin=*]
\item It assigns clear mathematical meaning (subobjects and slice diagrams in $(\E/S)$) to DoT components.
\item It formalizes synthesis via closure-based validation together with a variance-reversed information-colimit / slice-limit construction, with a precise split between the posetal and general-arrow settings (Theorem \ref{thm:summarization_colim_info}, Cor.~\ref{cor:general_sheafification}).
\item It demonstrates semantic invariance under isomorphic rearrangements (Proposition \ref{prop:robustness_universality}) and compositional gluing via pullbacks/fiber products of limits (Proposition \ref{prop:pasting_app}).
\item It structurally extends linear CoT in the posetal setting (Proposition \ref{prop:cot_special_case} and Theorem \ref{thm:separation}), matching intuitive gains from branching with a crisp categorical witness.
\end{itemize}

In summary, the primary advantages of DoT include an auditable reasoning trace, explicit compositional structure, and a clear theoretical target.

\vspace{5ex}
\bibliography{reference}
\bibliographystyle{plainnat}

\clearpage
\appendix
\renewcommand{\appendixpagename}{\centering \huge Appendix}
\appendixpage
\counterwithin{theorem}{section}

\startcontents[section]
\printcontents[section]{l}{1}{\setcounter{tocdepth}{2}}
\clearpage

\section{Worked Example Trace}
\noindent This example shows DoT applied to a problem with multiple independent conditions, demonstrating branching and invalidation.
\medskip

\noindent\texttt{@node id=1 role=problem}\\
\texttt{Is the point (4, 6) inside the region defined by x > 2, y > 5, and x < 10?}\\
\texttt{@node id=2 role=proposer}\\
\texttt{@edge src=1 dst=2 kind=use}\\
\texttt{@prop id=2 \{ "gt": [4, 2] \}}\\
\texttt{First, I will check the condition x > 2. For the point (4, 6), x=4. The inequality is 4 > 2, which is true.}\\
\texttt{@node id=3 role=critic}\\
\texttt{@edge src=2 dst=3 kind=critique}\\
\texttt{The arithmetic is correct. This condition holds.}\\
\texttt{@status target=2 mark=validated}\\
\texttt{@node id=4 role=proposer}\\
\texttt{@edge src=1 dst=4 kind=use}\\
\texttt{@prop id=4 \{ "gt": [6, 5] \}}\\
\texttt{Next, I will check the condition y > 5. For the point (4, 6), y=6. The inequality is 6 > 5, which is also true.}\\
\texttt{@node id=5 role=critic}\\
\texttt{@edge src=4 dst=5 kind=critique}\\
\texttt{Correct. The second condition holds.}\\
\texttt{@status target=4 mark=validated}\\
\texttt{@node id=6 role=proposer}\\
\texttt{@edge src=1 dst=6 kind=use}\\
\texttt{@prop id=6 \{ "gt": [6, 0] \}}\\
\texttt{Let me try an alternative, irrelevant check. Maybe the condition was y > 0. For y=6, 6 > 0 is true.}\\
\texttt{@node id=7 role=critic}\\
\texttt{@edge src=6 dst=7 kind=critique}\\
\texttt{This proposition is arithmetically correct, but it addresses a condition not present in the problem statement. This path is irrelevant.}\\
\texttt{@status target=6 mark=invalidated}\\
\texttt{@node id=8 role=proposer}\\
\texttt{@edge src=1 dst=8 kind=use}\\
\texttt{@prop id=8 \{ "lt": [4, 10] \}}\\
\texttt{Finally, I check the third condition, x < 10. For x=4, 4 < 10 is true.}\\
\texttt{@node id=9 role=critic}\\
\texttt{@edge src=8 dst=9 kind=critique}\\
\texttt{This is correct. The final condition is met.}\\
\texttt{@status target=8 mark=validated}\\
\texttt{@node id=10 role=summarizer}\\
\texttt{@edge src=2 dst=10 kind=use}\\
\texttt{@edge src=4 dst=10 kind=use}\\
\texttt{@edge src=8 dst=10 kind=use}\\
\texttt{Summary: All three conditions (x > 2, y > 5, and x < 10) are met for the point (4, 6). Therefore, the point is inside the specified region. The validated propositions are ID=2, ID=4, and ID=8.}

\section{DoT Serialization and Validation Details}
\label{app:serialization}

This section provides the complete details for the typed serialization, validation, and extraction mechanisms summarized in Section \ref{subsec:serialization_extraction}.

\paragraph{Serializer.}
We present a node-first, into-the-current-node serializer that guarantees acyclicity of the operational provenance DAG and supports online checks. The serializer is optional for usability, but recommended whenever formal guarantees or auditability are invoked. When it is disabled, DoT traces remain interpretable as natural language; the formal guarantees in Section~\ref{sec:topos_formalization} apply to the typed subtrace (if any), and free text is semantically inert.

\subsection{Grammar and Record Specification}

\paragraph{Node identifiers and roles.}
Each node carries a fresh natural-number ID and an explicit role:
\[
\texttt{@node id=}\langle n\rangle\texttt{ role=}\{\texttt{problem,proposer,critic,summarizer}\}.
\]
IDs are strictly increasing in emission order.

\paragraph{Typed edges and emission order.}
Edges are explicit, typed dependencies into the current node $j$:
\[
\texttt{@edge src=}\langle i\rangle\ \texttt{dst=}j\ \texttt{kind=}\{\texttt{refine,critique,use}\},\quad i<j.
\]
Well-formedness requires that sources \texttt{src} refer only to previously emitted node IDs. For critiques, the dependency direction is proposer $\to$ critic (the critic depends on the proposition it evaluates).

\paragraph{Admissible role/kind pairs (type table).}
Let $\mathrm{r}(\cdot)$ be node roles. Allowed triples for an \texttt{@edge src=i dst=j kind=k} are:
\begin{itemize}[leftmargin=1.2em,itemsep=2pt]
\item \texttt{kind=critique}: $\mathrm{r}(i)=\texttt{proposer}$, $\mathrm{r}(j)=\texttt{critic}$. A critic node's block may only contain edges of this kind.
\item \texttt{kind=refine}: $\mathrm{r}(i)\in\{\texttt{proposer},\texttt{critic}\}$, $\mathrm{r}(j)=\texttt{proposer}$. This edge is procedural unless the validator also accepts a typed strengthening witness; repair after invalidation is not assumed to entail the rejected proposition. If both endpoints are proposers and the witness certifies that the new proposer $j$ strengthens the old proposer $i$, extraction adds the semantic arrow $j\to i$, opposite to the operational edge direction.
\item \texttt{kind=use}: $\mathrm{r}(i)\in\{\texttt{problem},\texttt{proposer}\}$, $\mathrm{r}(j)\in\{\texttt{proposer},\texttt{summarizer}\}$. A \texttt{problem} source is contextual and is not itself included in the validated-proposition meet. Critics are relational and cannot be a source for \texttt{use} edges; instead, their semantic effect is reified via certified \texttt{@entails} or \texttt{@eq} records.
\end{itemize}

\paragraph{Arrow declarations (entailments) between proposers.}
Critiques can declare typed entailments between proposer nodes that become arrows in the extracted diagram:
\[
\texttt{@entails src=}\langle i\rangle\ \texttt{dst=}\langle k\rangle\ [\texttt{witness=}\langle j\rangle],
\]
with typing requirements: $\mathrm{r}(i)=\mathrm{r}(k)=\texttt{proposer}$, and the \texttt{witness}, if written, must be a \texttt{critic} node $j$ whose typed certificate is accepted by the validator. If the witness field is omitted, the current node must be a critic and is used as the witness. The intended meaning is $P_i\Rightarrow P_k$, hence a morphism $D_G(i)\to D_G(k)$ over $S$ (an inclusion $P_i\le P_k$ in the predicate fragment). Entailments are accepted only when certified by the typed witness and, in solver-backed fragments, by the corresponding entailment check.

\paragraph{Equivalence declarations.}
The basic syntax
\[
\texttt{@eq src=}\langle i\rangle\ \texttt{dst=}\langle k\rangle\ [\texttt{witness=}\langle j\rangle]
\]
is an abbreviation for mutual entailment in the predicate/posetal fragment. In the general slice fragment, it is accepted only when the witness certifies an isomorphism over $S$ or explicit inverse/path-equality data. It is not, by itself, a general path-equality language. If one implements richer path equalities, those records are added to the same congruence used in Def.~\ref{def:index_cat}.

\paragraph{Validation marks.}
Critiques emit a status for their target proposition:
\[
\texttt{@status target=}\langle k\rangle\ \texttt{mark=}\{\texttt{validated,invalidated}\}\ [\texttt{just=}\langle i\rangle].
\]
Only propositions with a \texttt{validated} status can contribute objects to the categorical synthesis diagram, and only when selected by the summarizer. A status record is well formed only when emitted by a critic that has a \texttt{kind=critique} edge from the target, and the optional \texttt{just} field must name that critic. We adopt a monotone state discipline: a proposition's status may be set only once (first-writer wins).

\paragraph{Lexical discipline (fencing).}
Typed records must begin with the reserved sigil \texttt{@}. Free-form natural-language text lines must not begin with \texttt{@}. This ensures unambiguous lexing. To embed arbitrary text (including leading \texttt{@}) we use a length-prefixed fence that is streaming-safe:
\[
\texttt{@@len=}\langle N\rangle\texttt{@@}\ \ \underbrace{\text{$\langle$exactly $N$ bytes of raw text$\rangle$}}_{\text{may contain @ and newlines}}.
\]

\paragraph{Concrete grammar (BNF).} Let $\mathbb N$ be decimal naturals and $\mathsf{Role}$ be the set of roles.
\[
\begin{array}{rcl}
\mathsf{Trace} &::=& \mathsf{NodeBlock}^+ \\
\mathsf{NodeBlock} &::=& \mathsf{Node}\ ;\ \mathsf{BlockItem}^* \\
\mathsf{BlockItem} &::=& \mathsf{Edge}\ \mid\ \mathsf{Status}\ \mid\ \mathsf{Entails}\ \mid\ \mathsf{Eq}\ \mid\ \mathsf{PropBlock}\ \mid\ \mathsf{Free}\\
\mathsf{Free} &::=& \mathsf{Text}\ \mid\ \mathsf{Esc}\\[0.3ex]
\mathsf{Node} &::=& \texttt{@node id=}\mathbb N\ \texttt{ role=}(\texttt{problem}|\texttt{proposer}|\texttt{critic}|\texttt{summarizer})\\
\mathsf{Edge} &::=& \texttt{@edge src=}\mathbb N\ \texttt{ dst=}\mathbb N\ \texttt{ kind=}(\texttt{refine}|\texttt{critique}|\texttt{use})\\
\mathsf{Status} &::=& \texttt{@status target=}\mathbb N\ \texttt{ mark=}(\texttt{validated}|\texttt{invalidated})\ [\texttt{ just=}\mathbb N]\\
\mathsf{Entails} &::=& \texttt{@entails src=}\mathbb N\ \texttt{ dst=}\mathbb N\ [\texttt{ witness=}\mathbb N]\\
\mathsf{Eq} &::=& \texttt{@eq src=}\mathbb N\ \texttt{ dst=}\mathbb N\ [\texttt{ witness=}\mathbb N]\\
\mathsf{PropBlock} &::=& \texttt{@prop id=}\mathbb N\ \mathsf{Prop}\\
\mathsf{Prop} &::=& \text{solver-checkable typed formula in the selected background theory}\\
\mathsf{Text} &::=& \text{arbitrary natural-language line not starting with @}\\
\mathsf{Esc} &::=& \texttt{@@len=}\mathbb N\texttt{@@}\ \mathsf{Bytes}^{\{\mathbb N\}}
\end{array}
\]
The grammar permits free text before or after typed records inside a node block. The validator, not the BNF alone, enforces role-specific side conditions such as ``status records may only be emitted by critic blocks'' and ``summarizer use-edges may point only to validated proposer nodes, except for optional contextual edges from the problem node.''

\subsection{Validation and Extraction}

\paragraph{Well-formedness judgment.} We write $\Gamma \vdash \mathsf{Trace}\;\mathsf{ok}$, where the context $\Gamma=(\mathsf{seen},\mathsf{role},\mathsf{state},\mathsf{current})$ tracks: seen node IDs, each node’s role, node states, and the current node being emitted. Selected rules:
\begin{gather*}
\frac{n > \max(\mathsf{seen}(\Gamma)) \quad r \in \{\texttt{problem},\texttt{proposer},\texttt{critic},\texttt{summarizer}\}}
{\Gamma \vdash \texttt{@node id=}n\texttt{ role=}r \quad \mathsf{ok}} \quad (\textsc{Node-Intro}) \\
\frac{i \in \mathsf{seen}(\Gamma) \quad i < j \quad \texttt{dst}=j=\mathsf{current}(\Gamma) \quad (\mathrm{r}(i),\mathrm{r}(j),k)\ \text{admissible}}
{\Gamma \vdash \texttt{@edge src=}i\texttt{ dst=}j\texttt{ kind=}k \quad \mathsf{ok}} \quad (\textsc{Edge-Intro}) \\
\frac{k \in \mathsf{seen}(\Gamma) \quad \mathsf{role}(k)=\texttt{proposer} \quad \mathsf{state}(k)=\texttt{active} \quad \mathsf{role}(\mathsf{current}(\Gamma))=\texttt{critic} \quad k\in\mathsf{critiqueTargets}(\mathsf{current}(\Gamma))}
{\Gamma \vdash \texttt{@status target=}k\texttt{ mark=}m \quad \mathsf{ok}} \quad (\textsc{Status-Intro})
\end{gather*}

\paragraph{Online validation.}
We employ a lightweight online validator $V$ with finite control over record kinds and register-based side conditions: (i) a monotone counter for the next ID, (ii) hash maps for roles and states, (iii) a table of critique targets and accepted typed witnesses, and (iv) a congruence-closure structure for equalities. Given a candidate token, $V$ performs local checks (e.g., $\texttt{src}<\texttt{dst}=\texttt{current\_id}$) and, in typed fragments, calls the selected decision procedure for entailment/equality witnesses. At inference, this validator provides masks to the LLM decoder, ensuring only well-formed sequences are generated.

\paragraph{Extraction map.}
Given a well-formed trace $H$, the syntactic extraction map $\Phi_{\mathrm{syn}}(H)$ returns: (i) a finite typed DAG $G=(V,E)$; (ii) the validated proposer subgraph; (iii) for each summarizer node $s$, the selected synthesis presentation $\J_{\Sigma}(s)$ generated by validated proposer nodes with accepted \texttt{use} edges into $s$; and (iv) an index category $\J_G$ generated by certified strengthening/refinement arrows, certified \texttt{@entails} arrows, and certified equivalence/path-equality relations as in Def.~\ref{def:index_cat}. Operational \texttt{use}, \texttt{critique}, and uncertified repair edges remain in the provenance DAG but do not become semantic arrows. After fixing a semantic interpretation $[\![\cdot]\!]$ for typed proposition blocks and certified arrows, $\Phi_{[\![\cdot]\!]}(H)$ returns the corresponding diagram $D_G:\J_G\to(\E/S)$ and its selected subdiagrams $D_{\Sigma}(s)$.

\begin{theorem}[Totality and Determinism of Extraction]
\label{thm:extraction_total_deterministic}
Any well-formed trace $H$ satisfying the ID and edge-typing rules yields a unique typed DAG $G$, unique selected synthesis presentations $\J_{\Sigma}(s)$ for summarizer nodes, and a unique syntactic index category $\J_G$ under $\Phi_{\mathrm{syn}}$. After fixing a semantic interpretation $[\![\cdot]\!]$, it yields a unique diagram $D_G:\J_G\to(\E/S)$ and selected subdiagrams $D_{\Sigma}(s)$. Excluding solver-call time, extraction is linear up to dictionary operations in the absence of nontrivial equality/path-equality declarations; with union-find equivalence classes it runs in $O(|H|\,\alpha(|H|))$ time and $O(|H|)$ space. Validation adds the sum of the selected theory solver costs. When general path equalities are present, we maintain congruence-closure on arrows (e.g., via e-graphs); this is near-linear in typical traces but can be super-linear in the worst case, depending on the signature and saturation strategy.
\end{theorem}

\subsection{Operational Semantics and Meta-Theory}

\paragraph{Standing scope.} We treat DoT’s semantics as a normative target: the LLM approximates the finite-limit/information-colimit behavior, while $V$ ensures the typed fraction is well-formed and auditable.

\begin{proposition}[Relative soundness to typed subtrace]
Any conclusion in Section~\ref{sec:topos_formalization} is a function of the extracted typed diagram $\Phi(H)$ alone; untyped/free text does not affect the semantics.
\end{proposition}

We give selected small-step rules over states $(G,\sigma,H)$ where $G$ is the partial DAG, $\sigma$ the node-state map, and $H$ the emitted prefix.

\paragraph{Rules (sketch).}
$\textsc{Proposer-Intro}$: on a well-typed \texttt{@node} with role=\texttt{proposer}, extend $G$ with a vertex $v$, set $\sigma(v)=\mathsf{active}$.
$\textsc{Critic-Intro}$: on role=\texttt{critic}, add the critic node.
$\textsc{Validate}$: on \texttt{@status} with \texttt{mark=validated} for target $u$ where $\sigma(u)=\mathsf{active}$, set $\sigma(u)=\mathsf{validated}$.
$\textsc{Invalidate}$: on \texttt{@status} with \texttt{mark=invalidated} for target $u$ where $\sigma(u)=\mathsf{active}$, set $\sigma(u)=\mathsf{invalidated}$.

\begin{theorem}[Order-Invariance (semantic; typed-content invariant)]
\label{thm:order_invariance}
Let a well-posed trace be one that (i) obeys the NodeBlock grammar, and (ii) contains at most one \texttt{@status} record for each proposer node ID (single-assignment). Normalize typed records by resolving omitted \texttt{witness} and \texttt{just} fields to their enclosing critic block and by discarding free text. Consider two well-posed traces, $H_1$ and $H_2$, that contain the same multiset of normalized typed records (\texttt{@node}, \texttt{@edge}, \texttt{@status}, \texttt{@entails}, \texttt{@eq}) and induce the same dependency partial order on those records after quotienting by validated equalities. Then their extracted index categories are isomorphic, $\J_{G_1} \cong \J_{G_2}$. For corresponding summarizer nodes, the selected synthesis presentations are also isomorphic. In the posetal case, the resulting subobjects are equal; in the general case, their reflected slice-limits are isomorphic.
\end{theorem}

\subsection{Algorithm and Implementation Details}

The state of the reasoning process (the DAG) is implicitly encoded within the auto-regressive history $H_t$. The LLM conditions its prediction on this history, using its attention mechanism to represent the current state of the DoT graph. Algorithm \ref{alg:dot_app} provides a high-level sketch.

\begin{algorithm}[ht!]
\caption{Diagram of Thought (DoT) Generation Process}
\label{alg:dot_app}
\begin{algorithmic}[1]
\State \textbf{Input:} Problem statement $\mathcal{P}$
\State Initialize generation history $H$ with serialized problem statement for node $v_1$; set current node $j \leftarrow 1$.
\State Initialize node states (e.g., in a dictionary) $\sigma[v_1] \leftarrow \text{initial}$.
\While{termination condition not met (e.g., max length, \texttt{<summarizer>} generated)}
\State Predict next role token $r \in T_{\text{roles}}$ based on history $H$: $r \sim \Lang(H)$.
\State Append $r$ to $H$.
\If{$r = \texttt{<proposer>}$}
\State Emit \texttt{@node id=} $j{+}1$ \texttt{ role=proposer}; set $j\leftarrow j{+}1$.
\State Emit zero or more well-typed \texttt{@edge src=} $i$ \texttt{ dst=} $j$ with $i<j$.
\State Generate proposition text $P_j$. Append records and text to $H$.
\State Update state: $\sigma[j] \leftarrow \text{active}$.
\ElsIf{$r = \texttt{<critic>}$}
\State Emit \texttt{@node id=} $j{+}1$ \texttt{ role=critic}; set $j\leftarrow j{+}1$.
\State Emit one or more \texttt{@edge src=} $k_m$ \texttt{ dst=} $j$ \texttt{ kind=critique}.
\State Generate critique text $C_j$; then emit one or more \texttt{@status target=} $k$ \texttt{ mark=validated|invalidated}.
\State Optionally emit certified \texttt{@entails} or \texttt{@eq} records.
\State Append records and text to $H$.
\State Update target states monotonically: if $\sigma[k]=\text{active}$, set $\sigma[k] \leftarrow m$.
\ElsIf{$r = \texttt{<summarizer>}$}
\State Emit \texttt{@node id=} $j{+}1$ \texttt{ role=summarizer}; set $j\leftarrow j{+}1$.
\State Emit zero or more \texttt{@edge src=} $i$ \texttt{ dst=} $j$ \texttt{ kind=use} from selected validated proposer nodes (and optionally the problem node as context).
\State Generate summary text $S$. Append records and text to $H$.
\State Set termination condition to true.
\EndIf
\EndWhile
\State \textbf{Output:} Final text $S$ from the summarizer node $v_S$.
\end{algorithmic}
\end{algorithm}

\paragraph{Controller-light decoding.} Decoding uses a deterministic, state-dependent mask derived from the validator's finite control, registers, and local typing maps. Illegal token classes are never sampled. There is no branching search or external resampling loop. The only state external to the LM is the validator's finite map store and any solver state needed for the chosen typed fragment.

\paragraph{Auxiliary supervision for summaries-as-limits.}
To align the \texttt{<summarizer>} with the normative target, one can add an auxiliary loss that penalizes violations of literals that are entailed by the finite meet of selected validated propositions. This complements the standard maximum-likelihood training on full traces.

\section{Additional Formal Details}

\subsection{From Critique Schemas to a Nucleus}
\label{subsec:schemas_to_nucleus_app}
Critique schemas do not automatically determine a Lawvere--Tierney topology. To avoid this gap, the main text assumes a fixed topology $j$ and its universal closure operator. When one wants to derive such a topology from typed critique schemas, only the modality-producing part of the schema should be compiled into pullback-stable closure sequents; certified entailments, equivalences, and exclusions remain separate typed records unless explicitly encoded as such sequents. Thus the items below are constraints on a topology, not arbitrary closure rules on the single lattice $\Sub(S)$:
\begin{itemize}[parsep=1pt, itemsep=1pt, topsep=0pt]
\item \textbf{Certified entailment} ($\mathsf{LE}$): a validated critique may introduce a morphism $P\to Q$ over $S$ (an inclusion $P\le Q$ in the predicate fragment). This is a semantic arrow in $\J_G$; it is not, by itself, a closure axiom.
\item \textbf{Modal validation sequent}: if a schema is intended to enlarge the closure of a predicate, it is compiled as a pullback-stable lower-bound sequent $Q\le c_A(P)$ for suitable subobjects $P,Q\hookrightarrow A$.
\item \textbf{Equivalence identification} ($\mathsf{EQ}$): a validated critique emitting mutual entailments in the predicate fragment or an isomorphism/equivalence record between proposer objects in the general slice fragment. This is handled by certified arrows and congruence equations, not by closure alone.
\item \textbf{Type refinement} ($\mathsf{TR}$): narrows a subobject $P$ by pullback along a mono $R\hookrightarrow S$, producing $P\wedge R$ and a certified strengthening arrow $P\wedge R\to P$.
\end{itemize}
The local operators on a topos form a complete lattice. For the closure constraints used here, the relevant modality sequents are lower-bound closure requirements of the form $Q\le c_A(P)$, closed under pullback. Satisfaction of such sequents is stable under arbitrary meets of local operators under the usual lattice order; hence the satisfying topologies, when nonempty, have a least element. For purely lower-bound closure constraints the chaotic topology satisfies the constraints, so nonemptiness is automatic. Equality, isomorphism, and exclusion constraints are not generated by this lattice argument alone; they must be certified separately or compiled into stable closure sequents whose consistency has been checked. The associated universal closure operator is then extensive, monotone, idempotent, pullback-stable, and finite-meet preserving by construction. This is the limited sense in which critique schemas can induce a validation modality.

\begin{lemma}[Rule closure \& nucleus completion]
Let $\mathcal R$ be a set of pullback-stable lower-bound closure sequents $Q_r\le c_{A_r}(P_r)$ generated by the modality-producing part of the typed critique schemas. If at least one Lawvere--Tierney topology satisfies $\mathcal R$, and if satisfaction of $\mathcal R$ is closed under meets of local operators (as it is for these lower-bound sequents), let $j_{\mathcal R}$ be the meet of all satisfying topologies. Then $j_{\mathcal R}$ is the least topology satisfying $\mathcal R$. Its associated universal closure operator $c^{\mathcal R}=(c^{\mathcal R}_A)_{A\in\E}$ is extensive, monotone, idempotent, pullback-stable, and finite-meet preserving. In particular, $c^{\mathcal R}_S$ is a nucleus on $\Sub(S)$.
\end{lemma}

\subsection{Further Categorical Results}

\begin{lemma}[Slice reflection and finite-limit stability]
\label{lem:slice_reflection_bc_app}
Let $a_j:\E\to \mathrm{Sh}_j(\E)$ be sheafification, which is left exact. The induced functor on slices
\[
a^{/S}_j:(\E/S)\to(\mathrm{Sh}_j(\E)/a_j S),\qquad (X\to S)\mapsto(a_jX\to a_jS),
\]
is also left exact, and hence preserves finite limits. Moreover, for a mono $m:P\hookrightarrow S$, the inverse image in $\E$ of the mono $a_jm:a_jP\hookrightarrow a_jS$ along the unit $\eta_S:S\to a_jS$ represents the $j$-closure $c_S(P)\hookrightarrow S$. This is the stability property required for the synthesis construction in Cor.~\ref{cor:general_sheafification} (finite limits in $(\E/S)$ followed by $a^{/S}_j$, and comparison back to $(\E/S)$ by pullback along $\eta_S$).
\end{lemma}

\begin{proposition}[Gluing validated diagrams via pullbacks of limits]
\label{prop:pasting_app}
Let $D_1:\J_1\to(\E/S)$ and $D_2:\J_2\to(\E/S)$ be selected validated diagrams whose overlap is a common subdiagram $K:\J_K\to(\E/S)$. Then, in a presheaf topos, there is a canonical isomorphism
\[
\lim(D_1\cup_K D_2)\ \cong\ \lim(D_1)\ \times_{\lim(K)}\ \lim(D_2),
\]
whenever the displayed union is the pushout of finite index shapes along the common subshape and the two diagrams agree on $K$. Hence, after reflection, the overall summary satisfies
\[
\mathrm{Summary}(D_1\cup_K D_2)\ \cong\ a^{/S}_j\!\big(\lim(D_1)\ \times_{\lim(K)}\ \lim(D_2)\big).
\]
\end{proposition}
\begin{proof}[Proof sketch]
Giving a cone over the union diagram is equivalent to giving a cone over $D_1$ and a cone over $D_2$ whose restrictions to $K$ agree. The representing object for such compatible pairs of cones is the pullback of the two individual limits over the overlap limit. Finally, $a^{/S}_j$ preserves finite limits by Lem.~\ref{lem:slice_reflection_bc_app}.
\end{proof}

\subsection{Validation and Termination}
\label{subsec:decoding_termination_app}
We enforce well-formedness with the validator $V$ (\Cref{app:serialization}). In inference, decoding is constrained by $V$; violations are rejected before token commitment when they are locally detectable, and completed typed records are checked before acceptance. Termination is enforced operationally by a budget on node blocks/typed records together with the explicit \texttt{<summarizer:end>} token.

\noindent\textbf{Defeasible validation.}
For more complex, non-monotone reasoning, one may choose in advance a finite priority set $\rho\in\{0,\dots,R\}$ and an increasing chain of Lawvere--Tierney topologies, with associated nuclei $c^{\le \rho}$. A retraction step from rank $\rho$ to $\rho'\!<\rho$ replaces the active modality by $c^{\le \rho'}$ on affected objects. This optional extension lies outside the monotone core; the order-invariance result applies for a fixed active modality.

\section{Detailed Proofs}
\label{app:proofs}

This appendix provides detailed proofs for the theorems, propositions, and lemmas presented in the main text. We assume familiarity with basic concepts from category theory and topos theory, as found in references like \citep{maclane2012sheaves, johnstone2002sketches}.

\subsection{Proofs for Section 3}

\begin{proof}[Proof of Theorem~\ref{thm:extraction_total_deterministic}]
The proof proceeds by demonstrating totality, determinism, and analyzing the computational complexity.

The syntactic extraction map $\Phi_{\mathrm{syn}}$ is defined for any trace $H$ that is deemed well-formed by the validator $V$. Well-formedness ensures that every record in the trace can be unambiguously parsed and assigned a syntactic action.

Every \texttt{@node} record has a unique, strictly increasing ID.
Every \texttt{@edge} record refers to a \texttt{src} ID that has already been emitted and a \texttt{dst} ID corresponding to the current node, preventing forward references and cycles in the operational dependency graph of records.
Typed records for status, entailments, and equalities have their targets and roles checked for validity.

Since every syntactically valid record has a defined extraction action (e.g., add a node, add an edge, update a state, record a summarizer selection, record an entailment/equivalence), the extraction process is defined for the entire trace. Hence, $\Phi_{\mathrm{syn}}$ is total on the set of well-formed traces.

We must show that a given well-formed trace $H$ maps to a single, unique syntactic diagram.

The set of nodes $V$ in the extracted DAG $G$ is uniquely determined by the set of \texttt{@node} records.
The set of operational provenance edges $E$ is uniquely determined by the set of \texttt{@edge} records.
The selected synthesis presentations are uniquely determined by the accepted \texttt{@edge kind=use} records into each summarizer and by the validated states of their source proposer nodes.
The index category $\J_G$ is formed by taking the generated category on proposer nodes, certified strengthening/refinement arrows, and certified entailment arrows, then quotienting by the smallest congruence generated by certified equivalence/path-equality records. Operational provenance edges that lack a typed semantic certificate do not enter $\J_G$. The smallest congruence is unique.

After a semantic interpretation $[\![\cdot]\!]$ is fixed, the diagram functor $D_G: \J_G \to (\E/S)$ maps each object (proposer node) to its interpreted object over $S$ and each certified arrow to its interpreted morphism over $S$.

Because each syntactic step is a deterministic function of the input trace, the final syntactic output $(G,\J_G,\{\J_{\Sigma}(s)\}_s)$ is unique; with fixed $[\![\cdot]\!]$, the semantic diagram $D_G$ and selected subdiagrams $D_{\Sigma}(s)$ are unique as well.

Now, let us calculate the complexity:
\begin{itemize}
\item Parsing the trace $H$ is a single linear pass, $O(|H|)$.
\item Building the graph structure, selected synthesis presentations, and semantic generators involves processing each record once. Using hash maps to store node information (roles, states), this takes $O(|H|)$ time and space, excluding solver calls.
\item When only node equivalences are present, managing these with a union-find data structure takes $O(|H|\,\alpha(|H|))$ time, where $\alpha$ is the inverse Ackermann function.
\item Solver-backed validation adds the sum of the costs of the selected decision procedures for the typed witnesses.
\item When path equalities are introduced, a more complex congruence closure algorithm is needed. While algorithms like e-graphs perform with near-linear amortized time in practice, their worst-case complexity can be higher depending on the specific theory. We state the complexity parametrically in this case.
\end{itemize}
The stated complexity bounds follow.
\end{proof}

\begin{proof}[Proof of Theorem~\ref{thm:order_invariance}]
The core insight is that the extraction map $\Phi$ is sensitive only to the set of normalized typed records and their dependency partial order, not the specific linear sequence in which they appear, provided that sequence is a valid topological sort of the dependency graph.

Since $H_1$ and $H_2$ contain the same multiset of normalized typed records, they will result in the same set of proposer nodes, the same dependency edges between nodes, the same status assignments, the same selected summarizer sources, the same set of entailment records, and the same set of declared equivalences.
The normalization step resolves implicit fields such as omitted witnesses, so equality is being tested on explicit records rather than on surface syntax.

The well-formedness rules (\texttt{src} $<$ \texttt{dst}, etc.) ensure that any valid trace is a topological sort of the underlying operational dependency graph of records. If $H_1$ and $H_2$ induce the same dependency partial order, they are simply two different valid topological sorts of the same abstract structure.

The extraction map $\Phi$ constructs the diagram by first identifying all nodes, provenance edges, summarizer selections, and semantic generators from the records and then applying the validated equalities. This process does not depend on the linear order of emission, only on the final set of normalized records and their dependency constraints. Therefore, $\Phi(H_1)$ and $\Phi(H_2)$ produce isomorphic abstract graphs, the same validated equalities, and thus isomorphic index categories $\J_G$ and selected presentations $\J_{\Sigma}$.
By Proposition \ref{prop:robustness_universality}, isomorphic diagrams have isomorphic limits. After reflection, the resulting summaries are isomorphic. In the posetal sub-case where the summary is a specific subobject (the meet-plus-closure), the summaries are equal.
\end{proof}

\subsection{Proofs for Section 4}

\begin{theorem}[DoT Process as Diagram Construction]
\label{app:proof_diagram_construction_full}
A DoT reasoning process generating a well-formed typed DAG $G=(V,E)$, together with a fixed semantic interpretation of typed propositions and certified arrows, defines a functor (a diagram) $D_G: \J_G \to (\E/S)$ with:
\begin{itemize}[parsep=1pt, itemsep=1pt, topsep=0pt]
\item Each typed proposer node $v$ mapped to an object $D_G(v)\to S$ in the slice (a subobject $D_G(v)\hookrightarrow S$ in the predicate/mono fragment);
\item Accepted critic records supply certified arrows, isomorphisms, or path equalities between proposer objects; critic nodes are witnesses for these records rather than objects of $D_G$;
\item Each arrow $v\to u$ in $\J_G$ is mapped to a certified morphism over $S$, witnessing entailment or strengthening coherence (posetal case: an inclusion $D_G(v)\hookrightarrow D_G(u)$).
\end{itemize}
Functoriality ensures that composite certified dependencies (via paths modulo $\equiv$) are respected in the slice.
\end{theorem}
\begin{proof}
We construct the functor $D_G: \J_G \to (\E/S)$ by defining its action on objects and morphisms and verifying that it satisfies the functoriality axioms.

As per Definition~\ref{def:index_cat}, the extraction map $\Phi$ applied to the trace yields a DAG $G$ and an index category $\J_G$ with proposer nodes as objects and syntactically certified arrows as morphism generators, modulo certified equality constraints.

For each object $v \in \mathrm{Ob}(\J_G)$, Assumption~\ref{ass:interp} states that its content can be interpreted as an object over $S$ (a subobject of $S$ in the predicate/mono fragment). We define the action of $D_G$ on objects as this interpretation:
\[
D_G(v) := [\![\mathrm{content}(v)]\!] \to S .
\]
This defines an object in the slice category $(\E/S)$.

Now let $\alpha:v\to u$ be a generator in $\J_G$. If $\alpha$ is induced by a certified strengthening/refinement record, Assumption~\ref{ass:interp} assigns it a morphism $D_G(v)\to D_G(u)$ over $S$. If $\alpha$ is induced by \texttt{@entails src=v dst=u}, the certified entailment directly supplies the same kind of morphism. If $\alpha$ is part of a certified \texttt{@eq} record, then in the predicate fragment it is mutual entailment, while in the general slice fragment the witness supplies an isomorphism and inverse equations. We then extend $D_G$ from generators to arbitrary morphisms by composition.

Well-definedness with respect to $\equiv$ follows from Assumption~\ref{ass:coh}: if two generated composites are identified by validated equality records, then their interpreted composites in the slice are equal. Identities map to identities, and composition in $\J_G$ maps to composition in $(\E/S)$. Hence $D_G$ is a functor.
Thus, $D_G$ is a valid functor from the index category $\J_G$ to the slice category $(\E/S)$.
\end{proof}

\begin{theorem}[Summarization as finite meet-plus-closure in the reflective subposet]
\label{app:proof_summarization_posetal_full}
Assume every proposer selected for synthesis is interpreted as a $c$-closed subobject of $S$ and every certified semantic arrow between selected proposers is an inclusion, so the selected diagram $D_{\Sigma}$ lands in the thin category $\Sub(S)$. Let $V_{\Sigma}$ be the finite set of selected validated proposer nodes. Then the finite-limit summary is
\[
\mathrm{Summary}_{\Sigma}
=
\lim\nolimits_{\Sub(S)}D_{\Sigma}
=
\bigwedge_{v\in V_{\Sigma}}D_G(v),
\]
equivalently the finite colimit of $D_{\Sigma}^{\op}$ in $\Pred(S)=\Sub(S)^{\op}$. The reflected form
\[
c\Big(\bigwedge_{v\in V_{\Sigma}}D_G(v)\Big)
\]
is equal to this meet.
\end{theorem}
\begin{proof}
Work in $\Sub(S)$ ordered by inclusion, and in the opposite poset $\Pred(S)=\Sub(S)^{\op}$ (information order).
For a finite family $\{P_v\}$, the join (colimit) in $\Pred(S)$ corresponds to the meet (intersection) in $\Sub(S)$:
\[
\bigvee\nolimits_{\Pred(S)} P_v \;=\; \bigwedge\nolimits_{\Sub(S)} P_v.
\]
More generally, since $\Sub(S)$ is thin, the finite limit of the selected posetal diagram is the greatest lower bound of its objects. Equivalently, after passing to the opposite diagram $D_{\Sigma}^{\op}:\J_{\Sigma}^{\op}\to\Pred(S)$, this same object is the information-order colimit.
Now assume each validated proposition is $c$-closed, i.e.\ $c(P_v)=P_v$, and that $c$ is a nucleus (finite-meet preserving;
Assumption~\ref{ass:j-topology}). Then the meet of validated propositions is again $c$-closed:
\[
c\!\Big(\bigwedge_v P_v\Big)\;=\;\bigwedge_v c(P_v)\;=\;\bigwedge_v P_v.
\]
Thus, the synthesis object lies in the $c$-closed fragment and agrees with the information-colimit. We may write the summary as
$c(\bigwedge_v P_v)$ to match the general-arrow presentation, noting that under the standing assumptions the outer $c$ is redundant.
\end{proof}

\begin{corollary}[General case via slice sheafification]
\label{app:proof_summarization_general_full}
For a general selected diagram $D_{\Sigma}:\J_{\Sigma}\to(\E/S)$ with non-posetal arrows and certified coherence/path-equality relations already satisfied by the extracted functor, the synthesized summary is
\[
\mathrm{Summary}_{\Sigma}
\;\cong\;
\lim\nolimits_{\mathrm{Sh}_j(\E)/a_jS}\big(a^{/S}_j\circ D_{\Sigma}\big)
\;\cong\;
a^{/S}_j\!\big(\lim\nolimits_{\E/S} D_{\Sigma}\big),
\]
where $a^{/S}_j:(\E/S)\to(\mathrm{Sh}_j(\E)/a_j S)$ is the functor induced by sheafification.
\end{corollary}
\begin{proof}
The logic is analogous to the posetal case but lifted from posets to general categories, using finite limits rather than meets.

In the presheaf topos setting, the slice $(\E/S)$ is finitely complete. Since selected diagrams from finite runs have finite graph presentations, the required limit $L=\lim D_{\Sigma}$ is a finite-limit construction in $(\E/S)$: finite products collect cone components, and equalizers impose compatibility with the finitely many generating arrows. Certified path-equality records are checked when constructing the functor $D_{\Sigma}$; they are not used to repair an otherwise incoherent diagram at synthesis time. Intuitively, this limit enforces simultaneous satisfaction/compatibility of the selected validated constraints represented by the diagram.

The Lawvere--Tierney topology $j$ defines a reflective subcategory of $j$-sheaves, $\mathrm{Sh}_j(\E) \hookrightarrow \E$. The reflector is the sheafification functor $a_j$. This induces a left exact functor on slice categories, $a^{/S}_j: (\E/S) \to (\mathrm{Sh}_j(\E)/a_j S)$, as stated in Lemma \ref{lem:slice_reflection_bc_app}. This functor maps objects in the slice to their validated/sheaf counterparts over $a_jS$.

Since $a^{/S}_j$ is left exact, it preserves finite limits. Therefore, the summary (synthesis computed within the validated/sheaf slice) can be obtained by computing the finite limit in the ambient slice and then applying the reflector:
\[
\mathrm{Summary}_{\Sigma}
\cong
\lim_{\mathrm{Sh}_j(\E)/a_jS}(a^{/S}_j\circ D_{\Sigma})
\cong
a^{/S}_j(\lim_{(\E/S)} D_{\Sigma}).
\]

When restricted to subobjects, the sheafified mono lives over $a_jS$. Pulling it back along $\eta_S:S\to a_jS$ gives the $j$-closure $c_S$ of the original subobject. The colimit in the information order corresponds to the meet in inclusion order after reversing variance. Thus, for a posetal diagram, comparison back over $S$ gives $c_S(\bigwedge_v D_G(v))$, recovering the result of Theorem \ref{thm:summarization_colim_info}.
\end{proof}

\begin{theorem}[Conditional consistency via closure validation]
\label{app:proof_consistency_full}
Fix $\E=\Set^{\C^{\mathrm{op}}}$ and a closure $c$ on $\Sub(S)$. In the fixed-stage finite-family setting, if the selected validated family is jointly satisfiable relative to a fixed stage $a_0\in\C$, then the summary is non-initial. In the model-compact setting, finite satisfiability of the corresponding first-order family implies satisfiability in some model, but not necessarily in the intended/standard model and not necessarily as componentwise non-emptiness in a preassigned presheaf stage.
\end{theorem}
\begin{proof}
Let $L = \bigwedge_{v \in V_{\Sigma}} D_G(v)$ be the meet of the interpretations of the selected validated propositions. In the fixed-stage case, the summary is $\mathrm{Summary}_{\Sigma} = c(L)$. We show that the summary is a non-initial subobject of $S$.

The premise states that the finite family $\{D_G(v)\}_{v \in V_{\Sigma}}$ is jointly satisfiable relative to a fixed stage $a_0 \in \C$. In the presheaf topos $\E = \Set^{\C^{\op}}$, this means there exists an element $x \in S(a_0)$ such that for every $v \in V_{\Sigma}$, $x$ lies in the subset $D_G(v)(a_0) \subseteq S(a_0)$.

The existence of such an element $x$ implies that the intersection of these subsets is non-empty:
\[
\bigcap_{v \in V_{\Sigma}} D_G(v)(a_0) \neq \emptyset.
\]
In a presheaf topos, finite limits are computed componentwise. Therefore, the component of the meet subobject $L$ at stage $a_0$ is precisely this intersection:
\[
L(a_0) = \left(\bigwedge_{v \in V_{\Sigma}} D_G(v)\right)(a_0) = \bigcap_{v \in V_{\Sigma}} D_G(v)(a_0).
\]
Since this set is non-empty, the subobject $L$ is non-initial.

The closure operator $c$ is extensive, meaning that for any subobject $P$, we have an inclusion $P \hookrightarrow c(P)$. Since presheaf monomorphisms are componentwise injections, applying this to our meet $L$ at stage $a_0$ gives:
\[
L(a_0) \subseteq (c(L))(a_0).
\]
Since we established that $L(a_0)$ is non-empty, its superset $(c(L))(a_0)$ must also be non-empty.
The summary, $\mathrm{Summary}_{\Sigma} = c(L)$, has a non-empty component at stage $a_0$. Therefore, the summary is a non-initial subobject.

For the model-compact setting, compactness is applied to the first-order family represented by the typed propositions. If every finite subset is satisfiable together with the background theory, then there is a model and assignment satisfying the entire family. When $S$ and $c$ are interpreted in that same model, extensivity again preserves satisfiability after closure. This proves consistency in the model-theoretic sense while deliberately avoiding any intended-model or fixed-stage presheaf claim.
\end{proof}

\begin{proposition}[Robustness under Diagram Isomorphisms]
\label{app:proof_robustness_full}
Let $D_{\Sigma,1}: \J_{\Sigma,1} \to (\E/S)$ and $D_{\Sigma,2}: \J_{\Sigma,2} \to (\E/S)$ represent selected validated reasoning steps from two different runs. If there is an isomorphism of diagrams, then their slice-limits are isomorphic.
\end{proposition}
\begin{proof}
Let $(\sigma,\eta)$ be an isomorphism of diagrams, i.e.\ $\sigma:\J_1\to\J_2$ is an isomorphism of index categories and
$\eta:D_1 \Rightarrow D_2\circ \sigma$ is a natural isomorphism.
Precomposition with $\sigma$ induces an equivalence between cones over $D_2$ and cones over $D_2\circ\sigma$; moreover,
the natural isomorphism $\eta$ induces an equivalence between cones over $D_1$ and cones over $D_2\circ\sigma$.
Equivalences preserve terminal objects, so the terminal cone over $D_1$ (its limit) corresponds to the terminal cone over $D_2$
(its limit). Hence $\lim D_1 \cong \lim D_2$.
\end{proof}

\begin{proposition}[Properties of Synthesis]
\label{app:proof_synthesis_properties_full}
The synthesis operation, $\mathrm{Summary}(\mathcal{P}) = c(\bigwedge_{P \in \mathcal{P}} P)$, exhibits Monotonicity, Idempotence, and Conservativity.
\end{proposition}
\begin{proof}
Let $\mathcal{P} = \{P_v\}_{v\in V_{\Sigma}}$. The properties follow directly from the definition of the summary and the standard properties of meet ($\wedge$) in a poset and a closure operator ($c$).

For monotonicity, we want to show that $\mathrm{Summary}(\mathcal{P} \cup \{P_{\text{new}}\}) \subseteq \mathrm{Summary}(\mathcal{P})$.
The meet of a larger set of subobjects is always a subobject of the meet of a smaller set:
\[
\bigwedge_{P \in \mathcal{P} \cup \{P_{\text{new}}\}} P = \left(\bigwedge_{P \in \mathcal{P}} P\right) \wedge P_{\text{new}} \subseteq \bigwedge_{P \in \mathcal{P}} P.
\]
The closure operator $c$ is monotone: if $X \subseteq Y$, then $c(X) \subseteq c(Y)$. Applying $c$ to both sides of the inclusion above preserves the relation:
\[
c\left(\bigwedge_{P \in \mathcal{P} \cup \{P_{\text{new}}\}} P\right) \subseteq c\left(\bigwedge_{P \in \mathcal{P}} P\right).
\]
This is the desired result.

For idempotence, we want to show $\mathrm{Summary}(\mathcal{P}) = c( \bigwedge_{P \in \mathcal{P}} c(P) )$.
Using finite-meet preservation of $c$,
\[
c\left(\bigwedge_{P\in\mathcal P}c(P)\right)
=
\bigwedge_{P\in\mathcal P}c(c(P))
=
\bigwedge_{P\in\mathcal P}c(P).
\]
If all propositions in $\mathcal{P}$ are validated, then each is already $c$-closed, so $P=c(P)$, and the displayed expression equals $c(\bigwedge_{P\in\mathcal P}P)$.

For conservativity, assume $P_w \supseteq \bigwedge_{P \in \mathcal{P}\setminus\{P_w\}} P$. We want to show $\mathrm{Summary}(\mathcal{P}) = \mathrm{Summary}(\mathcal{P} \setminus \{P_w\})$.
The meet of all propositions in $\mathcal{P}$ is:
\[
\bigwedge_{P \in \mathcal{P}} P = \left(\bigwedge_{P \in \mathcal{P}\setminus\{P_w\}} P\right) \wedge P_w.
\]
Since $P_w$ contains the meet of the others, intersecting with $P_w$ does not change the result. That is, if $X \subseteq Y$, then $X \wedge Y = X$. Therefore:
\[
\left(\bigwedge_{P \in \mathcal{P}\setminus\{P_w\}} P\right) \wedge P_w = \bigwedge_{P \in \mathcal{P}\setminus\{P_w\}} P.
\]
Applying the closure operator $c$ to both sides gives:
\[
c\left(\bigwedge_{P \in \mathcal{P}} P\right) = c\left(\bigwedge_{P \in \mathcal{P}\setminus\{P_w\}} P\right),
\]
which is precisely $\mathrm{Summary}(\mathcal{P}) = \mathrm{Summary}(\mathcal{P} \setminus \{P_w\})$.
\end{proof}

\end{document}